\documentclass{article} 
\usepackage[preprint]{colm2025_conference}

\usepackage{microtype}
\usepackage{hyperref}
\usepackage{url}
\usepackage{booktabs}
\usepackage{lineno}

\usepackage{amsmath,amsfonts,bm}

\def\eqref#1{equation~\ref{#1}}

\def\1{\bm{1}}

\DeclareMathAlphabet{\mathsfit}{\encodingdefault}{\sfdefault}{m}{sl}
\SetMathAlphabet{\mathsfit}{bold}{\encodingdefault}{\sfdefault}{bx}{n}

\usepackage{tikz,lipsum,lmodern}
\usepackage[most]{tcolorbox}
\usepackage{hyperref}
\usepackage{url}
\usepackage{color,soul}
\usepackage{moresize}
\usepackage{graphicx}
\usepackage{algorithm}
\usepackage{algorithmic}
\usepackage{multirow}
\usepackage{enumitem}
\usepackage{tabularx}
\usepackage{booktabs}
\usepackage{pifont}
\usepackage{multicol}
\usepackage{float}
\usepackage[framemethod=default]{mdframed}
\usepackage{booktabs}
\usepackage{array}
\usepackage{inconsolata}
\usepackage{lmodern}

\DeclareUnicodeCharacter{03B7}{$\eta$}

\let\etaorig\eta%
\renewcommand{\eta}{\ensuremath{\etaorig}}

\title{Right Prediction, Wrong Reasoning: Uncovering LLM Misalignment in RA Disease Diagnosis}

\author{Umakanta Maharana\textsuperscript{1}, Sarthak Verma\textsuperscript{2}, Avarna Agarwal\textsuperscript{2}, Prakashini Mruthyunjaya\textsuperscript{2}, \\
\textbf{Dwarikanath Mahapatra\textsuperscript{3}, Sakir Ahmed\textsuperscript{2}, Murari Mandal\textsuperscript{1}} \\
\textsuperscript{1} RespAI Lab, KIIT Bhubaneswar, \\
\textsuperscript{2} KIMS Bhubaneswar, \textsuperscript{3} Monash University, Australia\\
\textbf{Correspondence:} \texttt{murari.mandalfcs@kiit.ac.in}
}

\begin{document}

\maketitle
\begin{abstract}
Large language models (LLMs) offer a promising pre-screening tool, improving early disease detection and providing enhanced healthcare access for underprivileged communities. The early diagnosis of various diseases continues to be a significant challenge in healthcare, primarily due to the nonspecific nature of early symptoms, the shortage of expert medical practitioners, and the need for prolonged clinical evaluations, all of which can delay treatment and adversely affect patient outcomes. With impressive accuracy in prediction across a range of diseases, LLMs have the potential to revolutionize clinical pre-screening and decision-making for various medical conditions. In this work, we study the diagnostic capability of LLMs for Rheumatoid Arthritis (RA) with real world patients data.
Patient data was collected alongside diagnoses from medical experts, and the performance of LLMs was evaluated in comparison to expert diagnoses for RA disease prediction. We notice an interesting pattern in disease diagnosis and find an unexpected \textit{misalignment between prediction and explanation}. We conduct a series of multi-round analyses using different LLM agents. The best-performing model accurately predicts rheumatoid arthritis (RA) diseases approximately 95\% of the time. However, when medical experts evaluated the reasoning generated by the model, they found that nearly 68\% of the reasoning was incorrect. This study highlights a clear misalignment between LLMs high prediction accuracy and its flawed reasoning, raising important questions about relying on LLM explanations in clinical settings. \textbf{LLMs provide incorrect reasoning to arrive at the correct answer for RA disease diagnosis.}
\end{abstract}


\section{Introduction}
Early disease detection plays a crucial role in improving patient outcomes, especially in areas with limited healthcare resources. Identifying diseases at early stages can help reduce the strain on healthcare systems and improve survival rates. However, early diagnosis is still a major challenge in clinical practice due to factors such as vague early symptoms, a shortage of medical professionals, and the lengthy nature of diagnostic procedures~\cite{celermajer2012cardiovascular,crosby2022early,heidari2011rheumatoid}. These challenges can delay treatment and negatively affect patient health. In this context, pre-screening tools that assist in the diagnostic process could offer a significant improvement by enabling faster, more efficient, and accurate disease identification.

Large Language Models (LLMs) have shown great promise as pre-screening tools in recent years, with their ability to analyze vast amounts of clinical data and assist in early diagnosis~\cite{gopeekrishnan2024accelerating,chang2024applications,cohen2024automatically,lucas2024reasoning}. Several studies have demonstrated that LLMs can perform with impressive accuracy in predicting a range of diseases, including cancer~\cite{hein2024learning}, cardiovascular conditions~\cite{han2023large}, and infectious diseases~\cite{omar2024utilizing}. Their ability to quickly process and synthesize information from patient data, such as medical records and clinical notes, positions them as a potential game-changer in healthcare, especially for underserved populations with limited access to expert medical practitioners. These models can provide preliminary insights, helping to triage patients, reduce waiting times, and enhance healthcare access in areas with a shortage of medical professionals.

However, while LLMs have shown strong \textit{predictive} performance, their \textit{reasoning} remains a critical area of concern. In clinical applications, the accuracy of a model's predictions is essential, but so is the transparency and trustworthiness of its reasoning. \textit{In particular, when LLMs provide incorrect explanations or justifications for their predictions, the consequences could be severe}. A correct diagnosis with flawed reasoning may lead to misinformed decisions about treatment or further testing, undermining the model's utility in clinical settings. The importance of aligning predictions with sound reasoning cannot be overstated, especially in healthcare, where errors in reasoning can have a direct impact on patient well-being. Therefore, it is crucial to assess not only the accuracy of LLM predictions but also the reliability and correctness of the explanations that underpin these predictions.

In this study, we investigate the diagnostic capabilities of LLMs for the prediction of Rheumatoid Arthritis (RA), a common autoimmune disease with complex and often subtle early symptoms. Our work explores the alignment between LLM predictions and their reasoning for RA diagnosis using real-world patient data. We compare the performance of LLMs to expert medical diagnoses, and we uncover a critical issue: despite the high prediction accuracy (approximately 95\%), the reasoning behind these predictions is often flawed. When the explanations generated by the LLMs were reviewed by medical experts, nearly 68\% of the reasoning tokens were found to be incorrect. This misalignment between high prediction accuracy and flawed reasoning raises significant concerns about the reliability of LLMs in clinical decision-making.

By verifying the reasoning with input from medical experts, our study reveals an important gap in the application of LLMs for disease prediction. While LLMs can accurately predict conditions like RA, their reasoning may not align with medical expertise, limiting their effectiveness as a trustworthy clinical tool. This finding emphasizes the need for further research to improve the interpretability and reliability of LLMs in healthcare settings, ensuring that they not only provide accurate predictions but also offer sound and medically valid reasoning to support those predictions.

\section{Related Works}
\textbf{Reasoning in LLMs.} Reasoning capabilities in LLMs have garnered significant attention in recent years, as researchers explore ways to enhance their ability to solve complex tasks~\cite{jaech2024openai,guo2025deepseek}. Early studies focused on the emergence of reasoning skills in LLMs, particularly through in-context and few-shot learning, where models demonstrated the ability to perform basic reasoning across a range of tasks~\cite{sun2023survey,kojima2022large,liu2023large}. Additionally, methods such as reinforcement learning (RL) and Monte Carlo Tree Search (MCTS) enable iterative reasoning over extended thought chains~\cite{li2025fastmcts,wu2024comparative,kumar2024training}, while feedback mechanisms like self-verification and error correction refine the reasoning process~\cite{guo2025deepseek,ma2025s}. Efforts to improve structured and cross-lingual reasoning further broaden LLMs’ applicability across languages and domains~\cite{qin2023cross}, with hybrid feedback systems ensuring real-time corrections and enhancing reasoning quality~\cite{behrouz2024titans}. 

\textbf{Rheumatoid Arthritis Diagnosis.} Rheumatoid arthritis (RA) affects approximately 5 out of every 1000 individuals~\cite{aletaha2018diagnosis} and has the potential to cause significant joint damage and disability. Early diagnosis and intervention plays a critical role in managing RA. Recent advancements in Natural Language Processing (NLP) have had significant impacts on the management and research of RA~\cite{benavent2024large,benavent2023using,venerito2023ai}.~\cite{humbert2023development}created an NLP system to extract RA-related outcomes from clinical notes, which achieved high accuracy in identifying key clinical features. Similarly, AI-based tools leveraging electronic health records (EHRs) have shown success in diagnosing RA and identifying related conditions like rheumatoid arthritis with interstitial lung disease (RA-ILD), with promising precision metrics~\cite{singhal2023large,o2021association,fukae2020convolutional,zhang2020diagnostic,Bird2022,Irfan2023}.

The use of LLMs, such as ChatGPT, in rheumatology is still in its early stages, with limited studies available on their effectiveness and shortcomings. Our work is the first systematic analysis of the popular LLMs (ChatGPT, Gemini, Mistral, Qwen) for diagnosing RA with real world patients data.

\begin{figure}[t]
    \centering
    \includegraphics[width=0.6\textwidth]{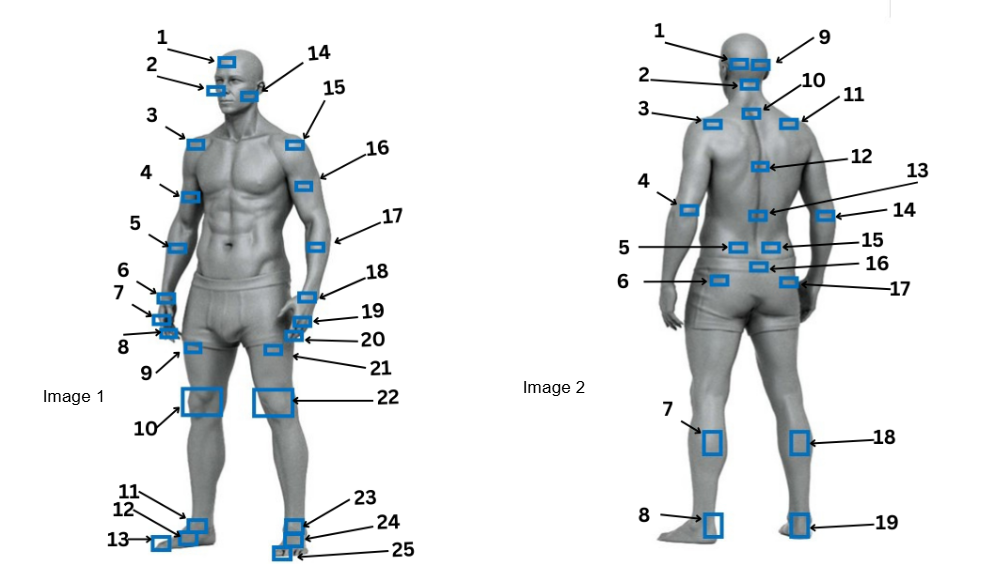}
    \caption{Diagram shown to the patient for indicating pain locations.}
    \label{fig:pain_locations}
\end{figure}

\begin{table}[t]
    \centering
    \begin{tabular}{c c}
        \toprule
        \textbf{Category} & \textbf{Description} \\
        \toprule
        Total patients           & 160 \\
        \midrule
        Gender distribution      & 15\% Male, 85\% Female \\
        \midrule
        RA diagnosis             & 85\% RA, 15\% Non-RA \\
        \midrule
        Languages used           & English, Odia \\
        \midrule
        Data collection method   & Online form \\
        \bottomrule
    \end{tabular}
    \caption{Key statistics of the PreRAID dataset}
    \label{tab:dataset_statistics}
\end{table}

\begin{table}[t]
\centering
\renewcommand{\arraystretch}{1.2}
\begin{tabular}{p{0.3\linewidth} p{0.6\linewidth}}
\toprule
\textbf{Patient information} & \textbf{Relevant fields in the online form} \\
\toprule
Demographic and Contact Details &
Timestamp, 
Email address,
First and last names,
Age,
Mother tongue,
Gender,
Mobile numbers \\
\midrule
Unique Identifiers and Geographic Data &
A unique KIMS ID for each patient \newline
Town/district information \newline
State information \\
\midrule
Symptomatology and Disease Progression &
Detailed responses on the primary problem faced by the patient \newline
Multiple entries for symptom onset (days, weeks, months, years) \newline
Comprehensive symptom checklists: pain in various body parts, early morning stiffness, joint deformities, and swelling \\
\midrule
Visual Aids for Symptom Localization &
Figure~\ref{fig:pain_locations} for marking specific pain locations \newline
Enhances precision of symptom reporting \\
\midrule
Additional Clinical and Lifestyle Information &
Presence of other symptoms: skin rashes, fever, mouth ulcers, ocular discomfort \newline
Impact on daily activities: sleep disturbances, difficulties in rising from a chair or bed, variations in pain with physical activity or rest \newline
Use and efficacy of painkillers \newline
Previous medication history for arthritis \\
\midrule
Follow-Up and Final Diagnosis &
Self-reported prescreening data \newline
Follow-up entries comprising doctor’s final diagnosis and explanatory notes \\
\bottomrule
\end{tabular}
\caption{Patient information collected through a structured online form. The details were filled by medical professionals in the presence of the patient.}
\label{tab:preraid}
\end{table}

\section{PreRAID Dataset}
\textbf{Data Collection and Description.} We present the PreRAID (Prescreening Rheumatoid Arthritis Information Database), a structured dataset developed to assess the diagnostic capabilities of large language models (LLMs) in the context of Rheumatoid Arthritis (RA) diagnosis. The dataset comprises 160 patient records, collected via a structured online form at the Kalinga Institute of Medical Sciences (KIMS), Bhubaneswar, India~\citep{pbmhPradyumnaMemorial}. Data gathering took place under the supervision of RA medical professionals, ensuring high-quality entries and proper patient consent throughout the process. The dataset captures a range of demographic details, including age, gender, and language~(refer Table \ref{tab:dataset_statistics}). Patients provided detailed descriptions of their primary complaints, with symptom onset recorded across varying time frames, ranging from days to years. Visual aids incorporated into the online form allowed patients to precisely indicate the location of pain, enhancing the accuracy of symptom reporting (refer Figure~\ref{fig:pain_locations}).

Beyond symptomatology, PreRAID includes associated symptoms such as skin rashes, fever, ocular discomfort, impact on daily activities, and medication history. The dataset also captures doctor-verified diagnoses and explanatory notes, providing a robust foundation for evaluating the reasoning behind LLM generated predictions. The dataset reflects real-world RA distribution: 85\% RA cases, 15\% non-RA cases, and 85\% female, 15\% male, a known clinical trend where RA predominantly affects women. Since we do not train an AI model but evaluate reasoning, this distribution does not impact the validity of our findings. Instead, PreRAID provides a controlled setting to assess how well LLMs justify their predictions, independent of dataset bias. The information collected for each patient is presented in Table~\ref{tab:preraid}

\textbf{Dataset Preprocessing.} The collected dataset is preprocessed and stored in a vector database for RAG-based analysis. \ding{182} \textit{Data Structuring.} Raw patient inputs are normalized into a standardized textual format, ensuring uniform representation across all records. \ding{183} \textit{Vectorization.} We embedded the structured text using GPT-4 text-embedding-3-large~\citet{openai2023gpt4embeddings}, a pre-trained embedding model that captures semantic relationships within patient data. These high-dimensional vector representations support effective similarity searches and context-aware reasoning. \ding{184} \textit{Storage in a Vector Database.}  The resulting embeddings are stored in a vector database, forming the knowledge backbone of our framework. This setup allows retrieval of patient records, enhancing diagnostic reasoning by improving contextual awareness. 

\begin{figure}[t]
    \centering
    \includegraphics[width=1.0\textwidth]{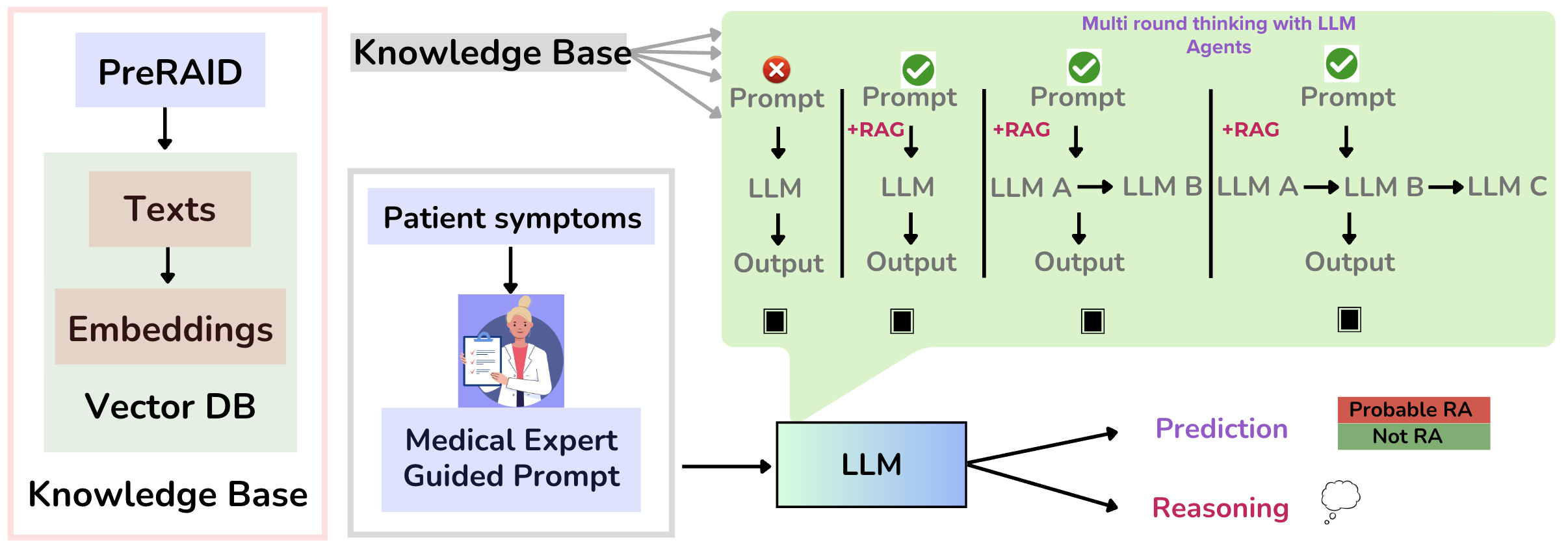}
    \caption{Overview of the framework for RA patients diagnosis. We conduct experiments with four different architectures: \textit{LLM without knowledge base (from PreRAID), LLM+knowledge base+RAG, 2 LLM Agents+knowledge base+RAG, 3 LLM Agents+knowledge base+RAG}. We store the final diagnosis and the reasoning tokens generated by each architecture.}
    \label{fig:knowledge}
\end{figure}

\section{RA Disease Diagnosis with LLMs}
\subsection{Experiment Settings}
We utilize the PreRAID dataset, which comprises 160 patient records, to conduct a series of experiments aimed at evaluating the diagnostic capabilities of large language models (LLMs). To examine the impact of varying knowledge base sizes on diagnostic performance, we construct knowledge bases of sizes: \{10, 20, 30, 40, 50, 60, 70, 80, 90\}. For each knowledge base size, we perform experiments using a range of LLMs, including GPT-3.5 Turbo, GPT-4o, GPT-4o-mini, Gemini 1.5 Flash, Gemini 2.0 Flash, and QWEN 2~\citep{geminiteam2023gemini, openai2023chatgpt, yang2024qwen2technicalreport}.

For each LLM, we test four distinct architectures: \ding{182}LLM without knowledge base ,\ding{183} LLM + Knowledge Base (from PreRAID) + RAG (retrieval-augmented generation), \ding{184} 2 LLM Agents + Knowledge Base + RAG, and \ding{185} 3 LLM Agents + Knowledge Base + RAG. This results in a total of 24 different configurations, which are used to generate diagnostic results for RA patients. 

To assess the effect of knowledge base size on performance, we incrementally expose the model to increasing amounts of historical patient data. Starting with a knowledge base of 10 records, the size is gradually increased in steps of 10, up to 90 records. For each configuration, we prompt the model to predict the diagnosis (either ``Rheumatoid Arthritis'' or ``Not Rheumatoid Arthritis'') and provide its reasoning. 

\subsection{Different LLM Architectures}
\textbf{LLM without knowledge base.} In the initial phase of our experiments, we evaluate a single LLM without any access to the knowledge base, which serves as a zero-shot predictor. The prompt for this configuration directs the model to analyze patient data and output the diagnosis based solely on its internal capabilities, without any additional contextual information. This setup provides a baseline for comparing performance against models that leverage external knowledge.

\begin{tcolorbox}[colback=blue!10, colframe=gray!20!gray, title=Prompt $\rightarrow$ \{LLM without knowledge base\}]
``Analyze the patient data thoroughly and then clearly state the diagnosis as `Rheumatoid Arthritis' or `Not Rheumatoid Arthritis'. Do not write any additional output or any patient information.\verb|\n|''
\end{tcolorbox}

\textbf{LLM + knowledge base + RAG.} We give access to historical diagnostic data (knowledge base) to the LLM which applies RAG and generate predictions accordingly. We use a similar prompt as before but include an instruction to incorporate the historical data.

\begin{tcolorbox}[colback=blue!10, colframe=gray!20!gray, title=Prompt $\rightarrow$ \{LLM + Knowledge Base + RAG\}]
``Analyze the patient data thoroughly and then clearly state the diagnosis as `Rheumatoid Arthritis' or `Not Rheumatoid Arthritis'. Do not write any additional output or any patient information. \verb|\n|. Use the provided data as historical diagnostic data : \textbf{``historical data''} \verb|\n|''
\end{tcolorbox}

\textbf{2 LLM Agents + knowledge base + RAG.} We introduce multi round thinking~\cite{tian2025think} by using two LLM agents. The architecture uses a two-round reasoning approach. In the first round, the agent analyzes the patient symptoms by comparing them with the historical data, and in the second round, it re-evaluates its initial response to produce a final diagnosis.

\begin{tcolorbox}[colback=blue!10, colframe=gray!20!gray, title=Prompt $\rightarrow$ \{2 LLM Agents + Knowledge Base + RAG\}]
\textbf{LLM Agent A}: ``Extract the patient symptoms from the provided user data {patient symptoms}. Extract the related symptoms and corresponding diagnosis from the knowledge base \textbf{``historical data''}. Compare them and provide diagnosis.''\\

\textbf{LLM Agent B}: ``Analyze this comparison {response of first agent} and then clearly state the diagnosis as `Rheumatoid Arthritis' or `Not Rheumatoid Arthritis'. Do not write any additional output or any patient information. \verb|\n|''
\end{tcolorbox}

\textbf{3 LLM Agents + Knowledge Base + RAG.} We create 3 LLM agents to  perform three rounds of reasoning. After an initial analysis and a subsequent verification of the first response, it conducts a final review before arriving at the diagnosis.

\begin{tcolorbox}[colback=blue!10, colframe=gray!20!gray, title=Prompt $\rightarrow$ \{3 LLM Agents + Knowledge Base + RAG\}]
\textbf{LLM Agent A}: ``Extract the patient symptoms from the provided user data {patient symptoms}. Extract the related symptoms and corresponding diagnosis from the knowledge base \textbf{``historical data''}. Compare them and provide diagnosis.''\\

\textbf{LLM Agent B}: ``{Response from first agent} Review this response is correct or incorrect and provide feedback by comparing it with the knowledge base \textbf{``historical data''}''.\\

\textbf{LLM Agent C}: ``Analyze this feedback {feedback} and then clearly state the diagnosis as `Rheumatoid Arthritis' or `Not Rheumatoid Arthritis'. Do not write any additional output or any patient information. \verb|\n|''
\end{tcolorbox}

\begin{figure}[t]
    \centering
    \includegraphics[width=1.0\textwidth]{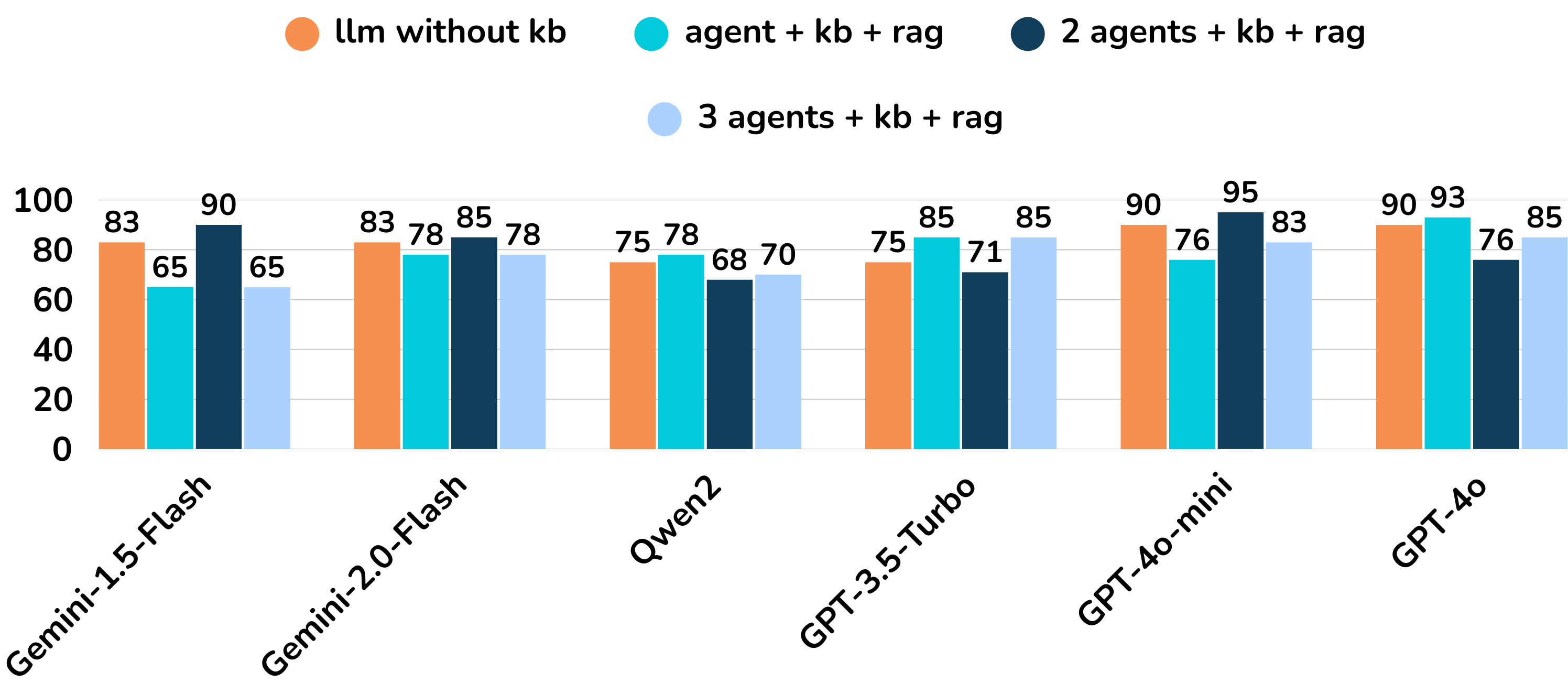}
    \caption{Accuracy of disease prediction across all the models by providing all the knowledge bases. Orange, sky blue, deep blue and light blue color represents the accuracy in llm without RAG, single agent+RAG, 2 agents+RAG, and 3 agents+RAG, respectively.}
    \label{fig:overall_accuracy}
\end{figure}

The experiments are conducted across all combinations: nine different knowledge base splits (ranging from 10 to 90 records), 6 LLM models, and 4 different architectural configurations. This setup enables a comprehensive evaluation of how both the amount of historical data and the iterative reasoning process affect the diagnostic accuracy of the predictions and the quality of the generated reasoning.

\begin{figure}[t]
    \centering
    \includegraphics[width=1.0\textwidth]{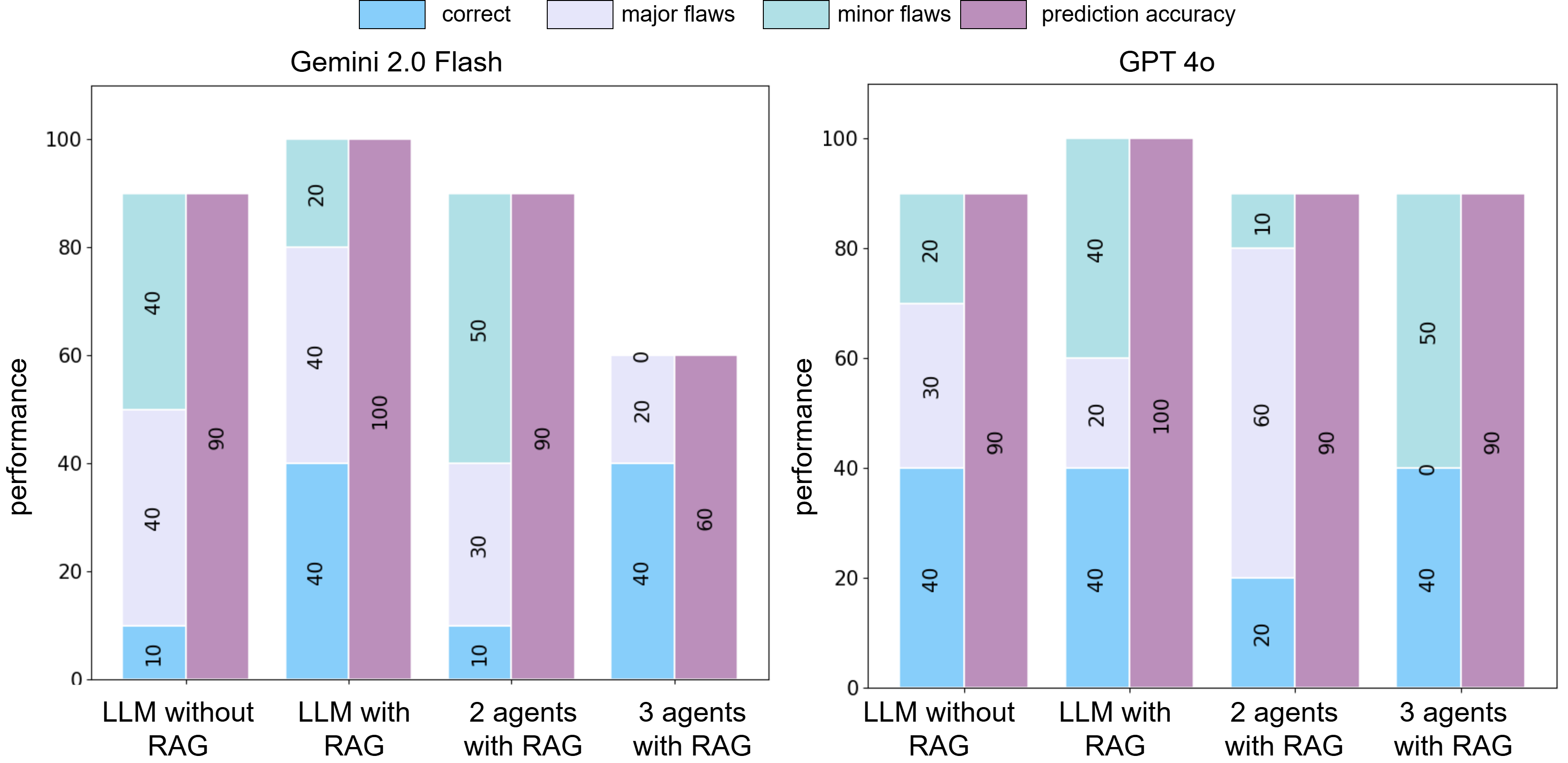}
    \caption{Comparison of diagnostic accuracy and expert-rated reasoning ratings for Gemini 2.0 Flash and GPT-4o across four architectures: LLM without RAG, LLM+RAG, 2 LLM agents+RAG, and 3 LLM agents+RAG}
    \label{fig:acc_vs_reason_four_methods_gemini_2.0_gpt-4o}
\end{figure}

\begin{figure}[t]
    \centering
    \includegraphics[width=1.0\textwidth]{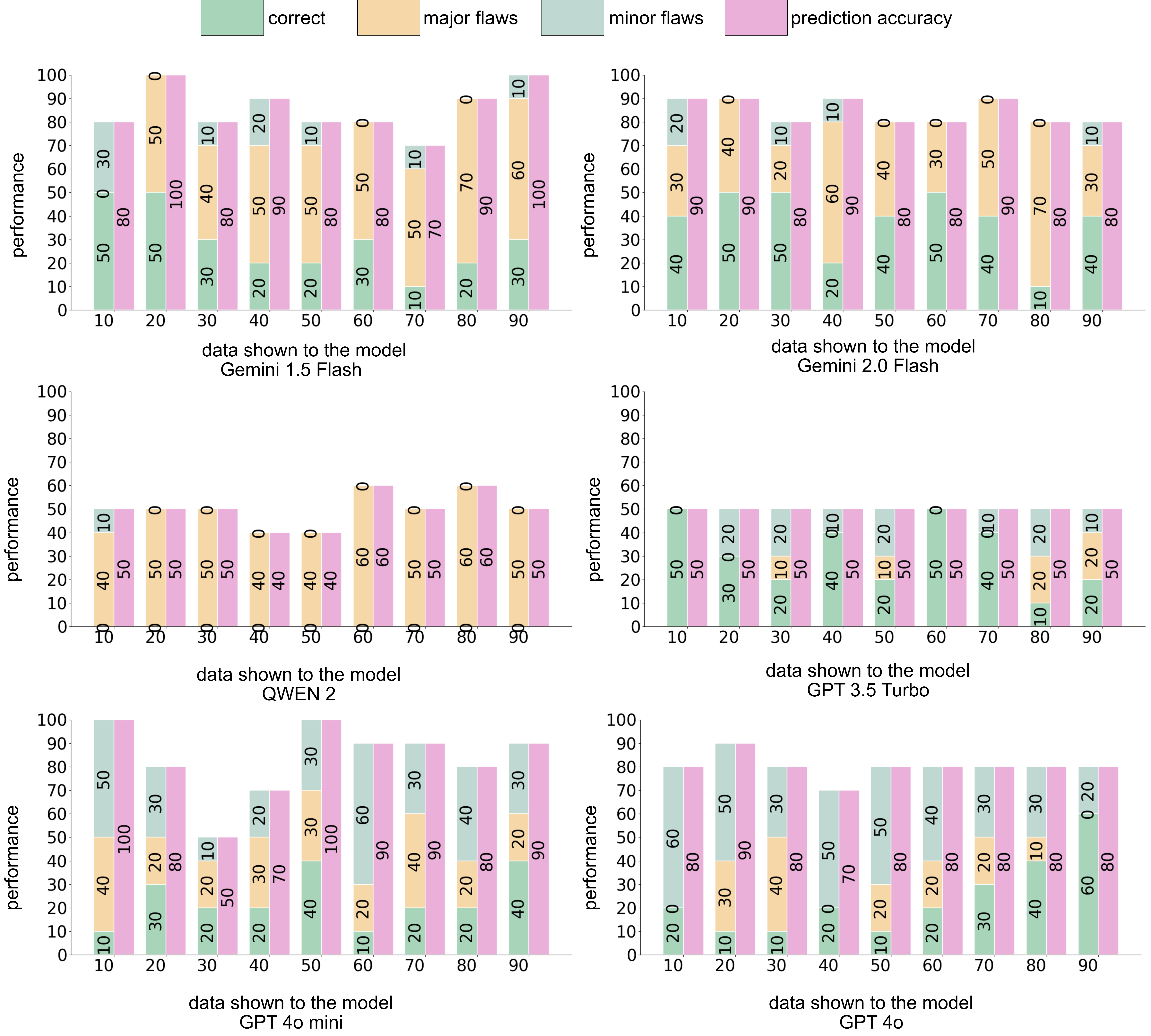}
    \caption{Reasoning performance across different LLM models in the 2 LLM agents+RAG setup. This figure compares the reasoning quality of various LLM models: GPT-3.5 Turbo, GPT-4o, GPT-4o-mini, Gemini 1.5 Flash, Gemini 2.0 Flash, and QWEN-2-7B, within the 2 LLM agents+RAG framework.}
    \label{fig:acc_vs_reasoning_2_llm_agents_all_models}
\end{figure}

\subsection{Results \& Discussion}
\textbf{Diagnostic accuracy analysis.} We have tried different combinations and most of the models are actually performing similar and providing high accuracy in disease prediction. We have tested 60 patients in all possible combinations. Figure 3 shows the prediction accuracy of all the models in different frameworks. We can see the orange color bars represent the accuracies of all the models in llm without knowledge base framework. This framework shows the highest accuracy as 90\% in GPT-4o and GPT-4o-mini models. And lowest in QWEN2. We can see the sky color bars represent the single agent with rag framework. It shows highest accuracy in GPT-4o model and lowest accuracy in Gemini-1.5-flash model. Gemini 2.0 flash shows 78\%, GPT-4o-mini 76\%, gpt3.5 turbo 85\% and qwen2 78\% accurate. We can see the deep blue color bars represent two agent framework with rag framework. It shows the highest accuracy in GPT4omini model and the lowest accuracy in qwen2 model. gemini1.5 flash shows 90\%, gemini2.0 flash shows 85\%, gpt3.5turbo models shows 71\%, gpt 4o shows 76\%. We can see that this framework with gpt 4o mini models shows the highest accuracy across all the combinations. We can see the light blue color represents three agents with rag framework. It shows highest accuracy in gpt 3.5 turbo and gpt 4o model and the lowest accuracy in gemini 1.5 flash model. gemini 2.0 flash shows 78\%, qwen2 shows 70\%, gpt4o mini shows 83\%.

\textbf{Reasoning validation data collected with RA experts.} Our experiments reveal that while all LLM-based architectures achieve over 90\% accuracy in diagnosing Rheumatoid Arthritis (RA), their underlying reasoning is not consistently aligned with expert clinical judgment. To investigate this discrepancy, we randomly selected 10 test patients: 5 diagnosed as RA and 5 as non-RA by ground truth and had domain experts evaluate the explanations provided by the LLMs for their diagnostic decisions. The RA experts classified the LLM reasoning into three categories: ``correct'', ``major flaw", and ``minor flaw''. A ``correct'' rating indicates that the LLM’s reasoning aligns with that of human experts, while a ``major flaw'' suggests that the model’s explanation is entirely unsound despite reaching the right diagnosis. A ``minor flaw' denotes partially accurate reasoning that still contains several errors.

Contrary to the assumption that high diagnostic accuracy implies sound reasoning, our findings expose a significant misalignment: LLMs often arrive at correct predictions through flawed or clinically unsound reasoning. This \textbf{“Right Prediction, Wrong Reasoning”} phenomenon raises critical questions about the interpretability and trustworthiness of LLMs in high-stakes medical applications. We present a comparative analysis of diagnostic accuracy and reasoning quality, underscoring the need for further research into improving the explanatory mechanisms of LLMs for reliable clinical deployment.  

\textbf{Reasoning analysis.} We conduct two types of analysis to compare the diagnostic accuracy and corresponding reasoning generated by the LLMs. 

\underline{\textit{Across four distinct architectures with Gemini 2.0 Flash and GPT4o.}} To assess the quality of LLM reasoning in RA diagnosis, we conducted experiments using the Gemini 2.0 Flash and GPT4o models with four distinct architectures: LLM without RAG, LLM+RAG, 2 LLM agents+RAG, and 3 LLM agents+RAG. We compared the diagnostic accuracy of each architecture against the reasoning ratings provided by RA experts (see Figure~\ref{fig:acc_vs_reason_four_methods_gemini_2.0_gpt-4o}). For Gemini 2.0 Flash, the prediction accuracies across the architectures were 90\%, 100\%, 90\%, and 60\%, while the percentage of instances rated as "correct" reasoning was 10\%, 40\%, 10\%, and 40\%, respectively. The gap in alignment is 80\%, 60\%, 80\%, and 20\%. Even when combining "correct" and "minor flaw" ratings, the gap between diagnostic accuracy and reasoning quality remained substantial: 40\%, 40\%, 30\%, and 20\%. This pronounced misalignment underscores a critical concern, particularly given the high-stakes nature of clinical applications.\par

Similarly, for GPT 4o, the prediction accuracy for different architectures are: 90\%, 100\%, 90\%, 90\% with corresponding ``correct'' reasoning as 40\%, 40\%, 20\%, 40\%, respectively. The gap in alignment is 50\%, 60\%, 70\%, and 50\%. Although, our sample size is small right now (total 10 test patients), the GPT 4o seems to perform slightly better than Gemini 2.0 Flash.This however is still far from the expected alignment between diagnosis accuracy and reasoning.\par

\underline{\textit{Across different LLM models in 2 LLM agents+RAG setup.}} We employed the 2 LLM agents+RAG architecture while varying the backend LLM models among GPT-3.5 Turbo, GPT-4o, GPT-4o-mini, Gemini 1.5 Flash, Gemini 2.0 Flash, and QWEN-2-7B~\citep{geminiteam2023gemini, openai2023chatgpt,yang2024qwen2technicalreport} (see Figure~\ref{fig:acc_vs_reasoning_2_llm_agents_all_models}). We observed a consistent pattern across models i.e., all models fail in proving correct reasoning for the high predictive accuracy for RA diagnosis. Notably, QWEN-2-7B failed to provide any correct reasoning, likely due to its significantly smaller model size. In contrast, all other models, which are proprietary and considerably larger, demonstrated better reasoning performance. 

\subsection{Final Takeaway} These are the final takeaways of our results:\\

\textbf{High Diagnostic Accuracy:} Across multiple architectures and models, LLMs consistently achieve over 90\% accuracy in predicting Rheumatoid Arthritis (RA), demonstrating strong potential as diagnostic tools.\par

\textbf{Misalignment in Reasoning:} Despite the high prediction accuracy, there is a significant gap between the diagnostic outcomes and the quality of the generated reasoning. Many LLMs arrive at correct predictions through explanations that are either partially or completely unsound, a phenomenon we term ``Right Prediction, Wrong Reasoning''. A similar observation has been reported in recent parallel studies (see, e.g.,~\cite{chenreasoning,yan2025recitation}). In our work, we demonstrate this phenomenon in the context of disease prediction by comparing real-world patient data and clinician diagnoses with the reasoning generated by LLMs.\par

\textbf{Architectural and Model Variations:} In experiments with Gemini 2.0 Flash and GPT-4o across four architectures, both models show high diagnostic performance, yet the reasoning quality remains inconsistent, with notable gaps (ranging from 20\% to 80\%) between correct predictions and valid reasoning. When comparing different LLM models under the 2 LLM agents+RAG framework, larger proprietary models (e.g., GPT-4o, GPT-3.5 Turbo) perform better in reasoning than smaller models like QWEN-2-7B, which fails to provide any correct reasoning.\par

\textbf{Implications for Clinical Use:} The pronounced misalignment between high prediction accuracy and flawed reasoning raises concerns about the interpretability and reliability of LLMs in clinical settings. For safe and effective deployment, it is crucial to improve the explanatory mechanisms so that model reasoning aligns closely with expert clinical judgment.\par

Overall, while LLMs show promise in early RA diagnosis, enhancing their reasoning capabilities is essential to ensure trustworthiness and clinical utility.

\section{Conclusion}
This study investigates the diagnostic performance of Large Language Models (LLMs) for Rheumatoid Arthritis (RA) using real-world patient data. While LLMs demonstrated high prediction accuracy (95\%) in diagnosing RA, their reasoning was often flawed, with nearly 68\% of explanations deemed incorrect by medical experts. This misalignment between accurate predictions and incorrect reasoning raises concerns about the reliability of LLMs in clinical decision-making. Although LLMs can assist in disease detection, their explanations must be improved to ensure clinical validity and patient safety. Our findings underscore the need for further research to enhance the interpretability and reasoning capabilities of LLMs, ensuring that they provide not only accurate predictions but also reliable, medically sound justifications for their diagnoses.

\section*{Acknowledgment}
This research is supported by the Science and Engineering Research Board (SERB), India under Grant SRG/2023/001686.


\bibliography{main}
\bibliographystyle{colm2025_conference}

\appendix
\newpage
\section{Appendix}

\textbf{Body Part Mapping to Text.} In Table~\ref{tab:mapping}, we describe the textual references for the body part selected by the patients, indicating pain locations in Figure~\ref{fig:pain_locations}. 

\begin{table}[!ht]
\caption{Body part mapping to corresponding text for Figure~\ref{fig:pain_locations}}
\label{tab:mapping}
\begin{tabular}{>{\bfseries}l l >{\bfseries}l l}
\toprule
\multicolumn{2}{c}{\textbf{Front Side}} & \multicolumn{2}{c}{\textbf{Back Side}} \\
\midrule
1 & Forehead & 1 & Suboccipital region \\
2 & Right TMJ joint & 2 & Posterior cervical region \\
3 & Right shoulder & 3 & Left shoulder (post) \\
4 & Right upper arm & 4 & Left elbow \\
5 & Right forearm & 5 & Lower back (left PSIS) \\
6 & Right wrist & 6 & Left hip buttock \\
7 & Right MCP (Metacarpal phalangeal joint) & 7 & Left calf \\
8 & Right PIP (Proximal interphalangeal joint) & 8 & Left heel \\
9 & Right thigh & 9 & Suboccipital region \\
10 & Right knee & 10 & Upper back \\
11 & Right ankle & 11 & Right shoulder (post) \\
12 & Right midfoot & 12 & Middle back \\
13 & Right toes & 13 & Lower midback \\
14 & Left TMJ joint & 14 & Right elbow \\
15 & Left shoulder & 15 & Right PSIS \\
16 & Left upper arm & 16 & Low back \\
17 & Left forearm & 17 & Right hip buttock \\
18 & Left writst & 18 & Right calf \\
19 & Right MCP (Metacarpal phalangeal joint) & 19 & Right heel \\
20 & Right PIP (Proximal interphalangeal joint) & & \\
21 & Right thigh & & \\
22 & Right knee & & \\
\bottomrule
\end{tabular}
\end{table}

\begin{figure}[!ht]
    \centering
    \includegraphics[width=0.8\textwidth]{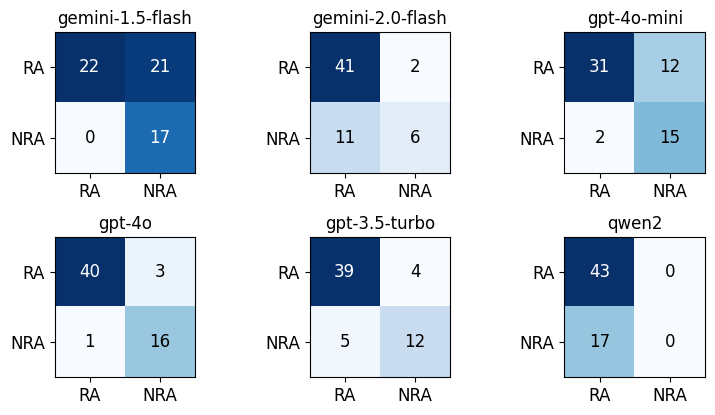}
    \caption{Confusion matrices for RA diagnosis in LLM without RAG}
    \label{fig:confusion_single_llm}
\end{figure}

\begin{figure}[!ht]
    \centering
    \includegraphics[width=0.8\textwidth]{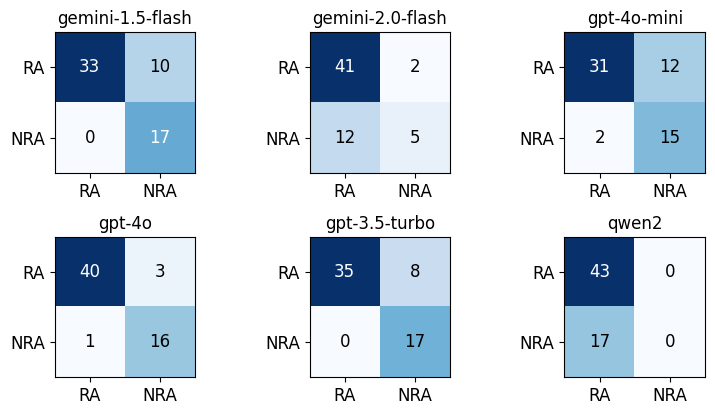}
    \caption{Confusion matrices for RA diagnosis in LLM+RAG}
    \label{fig:confusion_solo}
\end{figure}

\begin{figure}[!ht]
    \centering
    \includegraphics[width=0.8\textwidth]{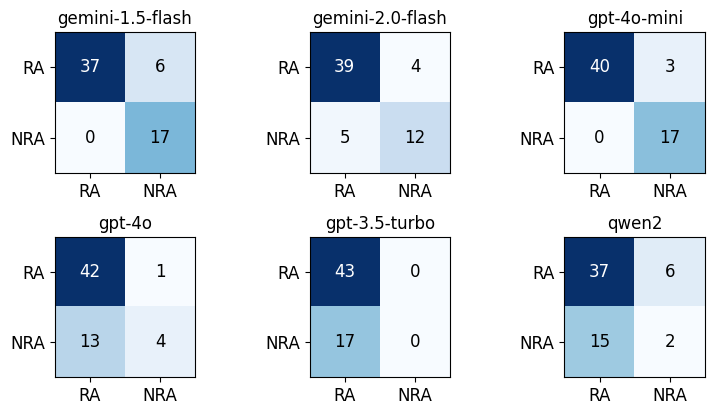}
    \caption{Confusion matrices for RA diagnosis in 2 LLM agents+RAG}
    \label{fig:confusion_duo}
\end{figure}

\begin{figure}[!ht]
    \centering
    \includegraphics[width=0.8\textwidth]{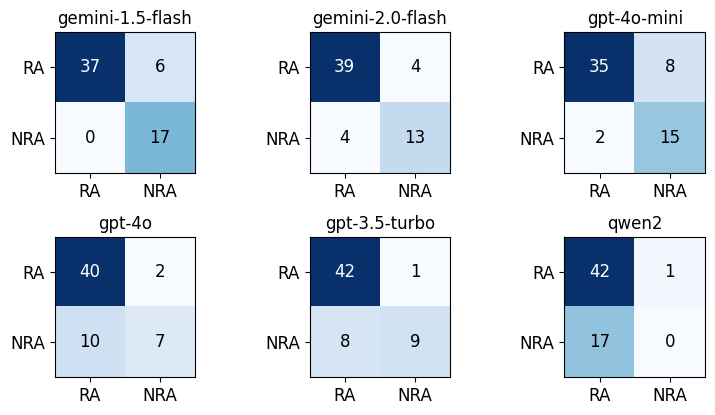}
    \caption{Confusion matrices for RA diagnosis in 3 LLM agents+RAG}
    \label{fig:confusion_trio}
\end{figure}

\textbf{Confusion Matrix Analysis.} We provide a comprehensive overview of the confusion matrices that assess the predictive performance of various LLM configurations in diagnosing rheumatoid arthritis (RA). These matrices capture the models' capabilities in accurately distinguishing between RA and non-RA (NRA) cases, although they do not reflect the soundness of the models' reasoning processes. In the single LLM without RAG configuration (Figure \ref{fig:confusion_single_llm}), high-capacity models such as GPT-4o and GPT-3.5 Turbo demonstrate better performance in predicting RA cases, while smaller models like QWEN-2-7B exhibit noticeable difficulties in classification. The LLM+RAG configuration (Figure \ref{fig:confusion_solo}) shows marked improvements, with models like Gemini-2.0 Flash significantly reducing misclassifications.

The 2 LLM agent+RAG (Figure \ref{fig:confusion_duo}) further enhances diagnostic accuracy, particularly for GPT-4o and GPT-3.5 Turbo, which now register minimal false classifications. However, some models, notably QWEN-2-7B, continue to face challenges in effectively discriminating between RA and NRA cases. In the 2 LLM agent+RAG configuration (Figure \ref{fig:confusion_trio}), most models, including GPT-4o and Gemini-2.0 Flash, attain near-optimal classification performance, underscoring the benefits of multi-agent collaboration in disease prediction.

It is crucial to emphasize that these confusion matrices exclusively quantify diagnostic accuracy. They do not assess the validity or depth of the models' underlying reasoning, which may remain imperfect even when classification metrics are high, as evidenced by expert evaluations.

\textbf{Prompt Engineering for Clinically Sound Reasoning.} 
To ensure that the LLM’s outputs not only identified the correct diagnosis but also articulated a medically valid rationale, our clinical team designed a bespoke prompt template. This instruction set guides the model to (1) reference specific patient findings, (2) weigh supporting and opposing evidence, and (3) explicitly rule out alternative diagnoses. By embedding these requirements directly into the prompt, we aimed to align the LLM’s explanatory process with the structured reasoning clinicians use in practice.

\scalebox{0.9}{
\begin{tcolorbox}[colback=red!10,colframe=gray!50!black,title=Prompt for Diagnostic Reasoning]
\small
\texttt{Patient information collected by survey: \{patient\_symptoms\}}\\
\texttt{Diagnosis: \{diagnosis\}}\\
\textcolor{blue}{\texttt{Provide reasons of diagnosis based on the following factors:
\begin{enumerate}
    \item Presence of early morning stiffness: higher is the duration of early morning stiffness, more is the chance of having an inflammatory arthritis
    \item Pain worsens after rest and improves with activity
    \item Involvement of the wrists, and small joints of the hands or feet makes it more likely however other peripheral joints may also be involved
    \item Good response to painkillers
    \item Additive distribution
    \item Gradual evolution of deformities
    \item Previous history of pain and swelling in specific joints as opposed to widespread swelling of body parts or swelling in all joints
    \item Absence of axial involvement or mid-foot involvement, especially in the first few years of the disease
    \item Definitive swelling in specific joints as opposed to widespread swelling of body parts or swelling in all joints
    \item Presence of skin rash is not seen in Rheumatoid arthritis and suggestive of other connective tissue disorders
    \item Low grade fever may be associated with Rheumatoid arthritis, but high-grade fever is unlikely
    \item Dryness of eyes and Dryness of mouth can be associated with Rheumatoid Arthritis
\end{enumerate}}}
- \texttt{Mention reasons in support of this diagnosis. \\
- Mention points against this diagnosis. \\
- Provide explanation for considering the diagnosis of Rheumatoid Arthritis and Not Rheumatoid Arthritis\\
- Also mention the other possible differential for that patient 
Do not output anything else.}
\end{tcolorbox}
}

\textbf{Examples of Flawed Reasoning Despite Correct Predictions.} As we show in the results, in several instances, the LLMs produce correct diagnostic outcomes yet demonstrated problematic reasoning in their justifications. For example, a model might accurately classify a case as rheumatoid arthritis (RA) while basing its decision on peripheral or outdated clinical indicators rather than the core pathophysiological evidence. In other instances, correct answers were supported by reasoning that relied on superficial statistical correlations or misinterpreted clinical data. These examples underscore the critical need to not only improve predictive accuracy but also to enhance the depth and reliability of the underlying reasoning processes in model-based diagnostics. Detailed examples of such cases are provided below.

\begin{tcolorbox}[colback=gray!5!white,colframe=blue!50!white,title= Patient Diagnosis with Gemini 2.0 Flash \{LLM with no RAG\} \{\textcolor{yellow}{Major Flaw}\}]
\begin{multicols}{2}
\begin{small}
\textbf{Patient Symptoms}\\

\textcolor{violet}{\textit{patient information:}}\\
- age: 45\\
- gender: female\\

\textcolor{violet}{\textit{problem description:}}\\
- primary problem: joint pain , joint swelling\\
- onset timing (in days): 0\\

\textcolor{violet}{\textit{symptoms and assessment:}}\\
- other symptoms: joint pain , joint swelling\\
- fever history: no\\
- joint pain: no\\
- swelling or deformity in joints: no\\
- redness in joints: no\\
- warmth in joints: no\\
- sleep disruption: no\\
- hours of sleep: 8\\
- effect of physical activity on pain: increase\\
- effect of rest on pain: reduce\\
- painkillers: no\\
- response to medication: not applicable\\
- skin rash: yes, hands / feet \\
- sunlight effect on rash: no\\
- grittiness in eyes: no\\
- eye dryness (use of eye drops): no\\
- difficulty swallowing dry foods: no\\
- difficulty sitting up: no\\
- difficulty getting up from lying position: no\\
- pain locations from image 1: right wrist, right mcp (metacarpal phalangeal joint), right pip (proximal interphalangeal joint)\\
- pain locations from image 2: no areas selected.\\
- previous history of pain and swelling in wrists, small joints of hands and feet : yes\\
- current medications: no\\

\textcolor{violet}{\textit{doctor's diagnosis:}}\\
- final diagnosis: \textcolor{red}{\textbf{rheumatoid arthritis}}\\
\end{small}
\columnbreak

\textbf{Reasoning by Gemini 2.0 flash}\\
\begin{tiny}
\textbf{The highlighted passages have been identified by our clinical reviewers as instances of flawed or misleading reasoning in the model’s justification for an RA diagnosis.}\\
\textcolor{violet}{\textit{Reasons in support of rheumatoid arthritis diagnosis:}}\\
\texttt{- joint pain and swelling: the patient presents with joint pain and swelling, which are primary symptoms of rheumatoid arthritis.\\
- pain locations: the pain locations (right wrist, right mcp, right pip) are consistent with ra, which often affects small joints of the hands and wrists.\\
\hl{- effect of physical activity and rest: pain increases with physical activity and reduces with rest.}\\
- previous history: the patient has a previous history of pain and swelling in the wrists and small joints of the hands and feet, and \hl{a prior diagnosis of rheumatoid arthritis.}\\
- age and gender: the patient is a 45-year-old female, which aligns with the typical demographic for ra onset.\\}

\textcolor{violet}{\textit{Points against rheumatoid arthritis diagnosis:}}\\
\texttt{- absence of fever: the patient has no fever history, while a low-grade fever can sometimes be associated with ra.\\
- absence of joint redness and warmth: the patient reports no redness or warmth in the joints, which are common inflammatory signs in ra.\\
- absence of sleep disruption: the patient reports 8 hours of sleep, indicating no sleep disruption, which is often present in ra patients.\\
- skin rash: the patient has a skin rash on hands and feet, which is not typical for rheumatoid arthritis and suggests other connective tissue disorders.\\
- painkillers: patient is not taking any painkillers.\\
- no swelling or deformity in joints: patient reports no swelling or deformity in joints.}\\

\textcolor{violet}{\textit{Explanation for considering the diagnosis}}:\\
\texttt{despite some contradictory symptoms, the patient's presentation strongly suggests rheumatoid arthritis due to the presence of joint pain, swelling, and specific joint involvement (wrists, mcp, pip). \hl{The previous history of ra} and the location of pain in the small joints of the hands and wrists are significant factors.}\\

\textcolor{violet}{\textit{Other possible differential for the patient}}:\\
\texttt{given the presence of a skin rash and the absence of some typical ra symptoms, other connective tissue disorders should be considered:\\
- systemic lupus erythematosus (sle): this autoimmune disease can present with joint pain, swelling, and skin rashes.\\
- psoriatic arthritis: this inflammatory arthritis can affect the joints and cause skin rashes (psoriasis).\\
- sjogren's syndrome: while primarily causing dryness, it can also present with joint pain and rashes.}\\

\end{tiny}
\end{multicols}
\end{tcolorbox}

\begin{tcolorbox}[colback=gray!5!white,colframe=blue!50!white,title= Patient Diagnosis with Gemini 2.0 Flash \{LLM with no RAG\} \{\textcolor{yellow}{Minor Flaw}\}]
\begin{multicols}{2}
\begin{small}
\textbf{Patient Symptoms}\\

\textcolor{violet}{\textit{patient information:}}\\
- age: 72\\
- gender: female\\

\textcolor{violet}{\textit{problem description:}}\\
- primary problem: joint pain , joint swelling , inability to move hand / feet\\
- onset timing (in days): 0\\

\textcolor{violet}{\textit{symptoms and assessment:}}\\
- other symptoms: joint pain , joint swelling , inability to move hand/feet, morning stiffness\\
- fever history: yes, less than 100f\\
- joint pain: no\\
- swelling or deformity in joints: no\\
- redness in joints: no\\
- warmth in joints: no\\
- sleep disruption: yes\\
- hours of sleep: 6\\
- effect of physical activity on pain: increase\\
- effect of rest on pain: increase\\
- painkillers: yes\\
- response to medication: moderate relief\\
- skin rash: yes, hands / feet\\
- sunlight effect on rash: no\\
- grittiness in eyes: itching\\
- eye dryness (use of eye drops): yes\\
- difficulty swallowing dry foods: no\\
- difficulty sitting up: yes\\
- difficulty getting up from lying position: yes\\
- pain locations from image 1: right shoulder, right upper arm, right wrist, right mcp (metacarpal phalangeal joint), right pip (proximal interphalangeal joint), right knee, right ankle, right midfoot, left shoulder, left upper arm, unknown, unknown, unknown, unknown, unknown, unknown\\
- pain locations from image 2: posterior cervical region, left shoulder (post), lower back (left psis), left hip buttock, upper back, right shoulder (post), right psis, unknown\\
- prior diagnoses: yes, rheumatoid arthraitis\\
- previous history of pain and swelling in wrists, small joints of hands and feet : yes\\
- current medications: yes, thyroid problem\\

\textcolor{violet}{\textit{doctor's diagnosis:}}\\
- final diagnosis: \textcolor{red}{\textbf{rheumatoid arthritis}}\\
\end{small}
\columnbreak

\textbf{Reasoning by Gemini 2.0 flash}\\
\begin{tiny}
\textbf{The highlighted passages have been identified by our clinical reviewers as instances of flawed or misleading reasoning in the model’s justification for an RA diagnosis.}\\
\textcolor{violet}{\textit{Reasons in support of rheumatoid arthritis diagnosis:}}\\
\texttt{- joint pain and swelling: the patient presents with joint pain and swelling, which is a primary symptom of rheumatoid arthritis.\\
- inability to move hands/feet: this limitation in movement due to joint issues aligns with ra's impact on joint function.\\
- morning stiffness: the presence of morning stiffness is a significant indicator of inflammatory arthritis like ra.\\
- sleep disruption: sleep disturbances are common in ra patients due to pain and discomfort.\\
- \hl{effect of physical activity and rest: pain increasing with both physical activity and rest suggests an inflammatory process.}\\
- moderate relief from medication: the patient experiences moderate relief from medication, which is typical in ra management.\\
- pain locations: the pain locations in the right shoulder, upper arm, wrist, mcp, pip, and knee are consistent with the peripheral joint involvement seen in ra.\\
- previous history: \hl{a prior diagnosis of rheumatoid arthritis strongly supports the current diagnosis.}\\
- history of pain and swelling in wrists and small joints: this historical pattern is highly characteristic of ra.\\
- eye dryness: eye dryness can be associated with rheumatoid arthritis\\}

\textcolor{violet}{\textit{Points against rheumatoid arthritis diagnosis:}}\\
\texttt{- absence of joint pain and swelling: the patient's assessment indicates "no" to joint pain and swelling, which contradicts the primary problem description.\\
- absence of redness and warmth in joints: ra often presents with redness and warmth in the affected joints, which are absent in this patient.\\
- skin rash: the presence of a skin rash on the hands and feet is atypical for ra and more suggestive of other connective tissue disorders.\\
- grittiness in eyes: the presence of grittiness in eyes is atypical for ra and more suggestive of other connective tissue disorders.\\}

\textcolor{violet}{\textit{Explanation for considering the diagnosis}}:\\
\texttt{despite some contradictory information, \hl{the historical diagnosis of ra}, the presence of morning stiffness, sleep disruption, pain patterns, and response to medication point towards a potential ra diagnosis. the "no" to joint pain and swelling could be a reporting error or a misinterpretation during the assessment. the skin rash and grittiness in eyes should be investigated further to rule out other conditions.}\\

\textcolor{violet}{\textit{Other possible differential for the patient}}:\\
\texttt{other connective tissue disorders: given the skin rash and grittiness in eyes, conditions like lupus, psoriatic arthritis, or sjogren's syndrome should be considered.\\
- osteoarthritis: although less likely due to the inflammatory nature of the symptoms, osteoarthritis could be a differential, especially considering the patient's age.}\\

\end{tiny}
\end{multicols}
\end{tcolorbox}

\begin{tcolorbox}[colback=gray!5!white,colframe=blue!50!white,title= Patient Diagnosis with GPT 4o \{LLM with no RAG\} \{\textcolor{yellow}{Major Flaw}\}]
\begin{multicols}{2}
\begin{small}
\textbf{Patient Symptoms}\\

\textcolor{violet}{\textit{patient information:}}\\
- age: 45\\
- gender: female\\

\textcolor{violet}{\textit{problem description:}}\\
- primary problem: joint pain , joint swelling\\
- onset timing (in days): 0\\

\textcolor{violet}{\textit{symptoms and assessment:}}\\
- other symptoms: joint pain , joint swelling\\
- fever history: no\\
- joint pain: no\\
- swelling or deformity in joints: no\\
- redness in joints: no\\
- warmth in joints: no\\
- sleep disruption: no\\
- hours of sleep: 8\\
- effect of physical activity on pain: increase\\
- effect of rest on pain: reduce\\
- painkillers: no\\
- response to medication: not applicable\\
- skin rash: yes, hands / feet\\
- sunlight effect on rash: no\\
- grittiness in eyes: no\\
- eye dryness (use of eye drops): no\\
- difficulty swallowing dry foods: no\\
- difficulty sitting up: no\\
- difficulty getting up from lying position: no\\
- pain locations from image 1: right wrist, right mcp (metacarpal phalangeal joint), right pip (proximal interphalangeal joint)\\
- pain locations from image 2: no areas selected.\\
- prior diagnoses: yes, rheumatoid arthraitis\\
- previous history of pain and swelling in wrists, small joints of hands and feet : yes\\
- current medications: no\\

\textcolor{violet}{\textit{doctor's diagnosis:}}\\
- final diagnosis: \textcolor{red}{\textbf{rheumatoid arthritis}}\\
\end{small}
\columnbreak

\textbf{Reasoning by GPT 4o}\\
\begin{tiny}
\textbf{The highlighted passages have been identified by our clinical reviewers as instances of flawed or misleading reasoning in the model’s justification for an RA diagnosis.}\\
\textcolor{violet}{\textit{Reasons in support of rheumatoid arthritis diagnosis:}}\\
\texttt{1. previous history of joint pain: the patient has a \hl{prior diagnosis of rheumatoid arthritis} with a history of pain and swelling in wrists, and small joints of hands and feet.\\
2. joint involvement: the pain is in the right wrist, right mcp, and right pip, which are common sites affected by rheumatoid arthritis.\\
3. effect of physical activity: \hl{pain increases with physical activity and reduces with rest, which aligns with rheumatoid arthritis symptoms.}\\
4. additive distribution: the involvement of multiple small joints conforms to the pattern seen in rheumatoid arthritis.\\
5. no axial or midfoot involvement: the absence of axial or midfoot involvement supports rheumatoid arthritis, as these are uncommon sites for the disease in early stages.}\\

\textcolor{violet}{\textit{Points against rheumatoid arthritis diagnosis:}}\\
\texttt{1. absence of severe morning stiffness: there's no mention of early morning stiffness, which is a typical symptom of rheumatoid arthritis.\\
2. no definitive swelling: the symptoms lack definitive swelling or deformity in joints, typically seen in active rheumatoid arthritis.\\
3. presence of skin rash: the patient has a skin rash on hands/feet, which is not characteristic of rheumatoid arthritis and suggests another connective tissue disorder.\\
4. no fever history: \hl{the absence of any fever history could be atypical for active rheumatoid arthritis, where low-grade fever might be present.}\\
5. good response to painkillers: there's no mention of a response to painkillers, which is often observed in rheumatoid arthritis management.}\\

\textcolor{violet}{\textit{Explanation for considering the diagnosis}}:\\
\texttt{the diagnosis of rheumatoid arthritis is considered due to the patient's prior established diagnosis and specific joint involvement typical of the disease, including the wrists and small joints of hands. the absence of complete symptoms may be explained by variability in expression or previous treatment success.}\\

\textcolor{violet}{\textit{Other possible differential for the patient}}:\\
\texttt{1. psoriatic arthritis: the presence of a skin rash could suggest an alternative diagnosis such as psoriatic arthritis, where skin involvement is common.\\
2. lupus (systemic lupus erythematosus): this condition can present with joint pain, swelling, and skin rash, typically in a photosensitive pattern, which may fit with the current symptoms.\\
3. reactive arthritis: if coupled with a recent infection, reactive arthritis is a possibility, although less likely due to the chronicity.\\
4. sjogren's syndrome: although not evident here, it can coincide with other autoimmune disorders and cause joint symptoms.}\\

\end{tiny}
\end{multicols}
\end{tcolorbox}

\begin{tcolorbox}[colback=gray!5!white,colframe=blue!50!white,title= Patient Diagnosis with GPT 4o \{LLM with no RAG\} \{\textcolor{yellow}{Minor Flaw}\}]
\begin{multicols}{2}
\begin{small}
\textbf{Patient Symptoms}\\

\textcolor{violet}{\textit{patient information:}}\\
- age: 72\\
- gender: female\\

\textcolor{violet}{\textit{problem description:}}\\
- primary problem: joint pain , joint swelling , inability to move hand / feet\\
- onset timing (in days): 0\\

\textcolor{violet}{\textit{symptoms and assessment:}}\\
- other symptoms: joint pain , joint swelling , inability to move hand/feet, morning stiffness\\
- fever history: yes, less than 100f\\
- joint pain: no\\
- swelling or deformity in joints: no\\
- redness in joints: no\\
- warmth in joints: no\\
- sleep disruption: yes\\
- hours of sleep: 6\\
- effect of physical activity on pain: increase\\
- effect of rest on pain: increase\\
- painkillers: yes\\
- response to medication: moderate relief\\
- skin rash: yes, hands / feet\\
- sunlight effect on rash: no\\
- grittiness in eyes: itching\\
- eye dryness (use of eye drops): yes\\
- difficulty swallowing dry foods: no\\
- difficulty sitting up: yes\\
- difficulty getting up from lying position: yes\\
- pain locations from image 1: right shoulder, right upper arm, right wrist, right mcp, right pip, right knee, right ankle, right midfoot, left shoulder, left upper arm, unknown (×6)\\
- pain locations from image 2: posterior cervical region, left shoulder (post), lower back (left psis), left hip buttock, upper back, right shoulder (post), right psis, unknown\\
- previous history of pain and swelling in wrists, small joints of hands and feet : yes\\
- current medications: yes, thyroid problem\\

\textcolor{violet}{\textit{doctor's diagnosis:}}\\
- final diagnosis: \textcolor{red}{\textbf{rheumatoid arthritis}}\\
\end{small}
\columnbreak

\textbf{Reasoning by GPT 4o}\\
\begin{tiny}
\textbf{The highlighted passages have been identified by our clinical reviewers as instances of flawed or misleading reasoning in the model’s justification for an RA diagnosis.}\\
\textcolor{violet}{\textit{Reasons in support of rheumatoid arthritis diagnosis:}}\\
\texttt{1. morning stiffness: the patient experiences morning stiffness, which is a common symptom associated with inflammatory arthritis, including rheumatoid arthritis.\\
2. joint involvement: the patient exhibits pain in the wrists and small joints of the hands, which aligns with characteristic joint involvement in rheumatoid arthritis.\\
3. history of rheumatoid arthritis: the patient has a \hl{prior diagnosis of rheumatoid arthritis} and a history of pain and swelling in specific joints, which supports the continuity of this diagnosis.\\
4. effect of physical activity and rest on pain: the patient's pain increases with rest and physical activity, which is common in rheumatoid arthritis where morning stiffness eases with activity.\\
5. low-grade fever: the patient reports a \hl{low-grade fever history, which can be associated with rheumatoid arthritis.}\\
6. use of painkillers: the patient experiences moderate relief with painkillers, suggesting effective symptomatic treatment for rheumatoid arthritis.}\\

\textcolor{violet}{\textit{Points against rheumatoid arthritis diagnosis:}}\\
\texttt{1. presence of skin rash: the patient has a skin rash, which is not typical in rheumatoid arthritis and is suggestive of other connective tissue disorders.\\
2. eye dryness: eye dryness can occur in rheumatoid arthritis, but its presence along with skin rash might indicate other conditions like sjögren's syndrome or lupus.\\
3. joint swelling and pain consistency: there is inconsistency in the description of joint swelling and pain, occasionally indicating absence of signs like warmth, redness, or definitive swelling.}\\

\textcolor{violet}{\textit{Explanation for considering the diagnosis:}}\\
\texttt{the patient's symptoms, history, and \hl{previous diagnosis of rheumatoid arthritis}, combined with classic signs such as morning stiffness and joint involvement, support the continuation of this diagnosis. despite the presence of a skin rash and eye symptoms, these could be interpreted as complexity or multi-faceted presentation rather than ruling out rheumatoid arthritis as the primary diagnosis.}\\

\textcolor{violet}{\textit{Other possible differential for the patient:}}\\
\texttt{1. sjögren's syndrome: the presence of dry eyes and potentially dry mouth could indicate sjögren's syndrome, especially if accompanied by joint symptoms.\\
2. systemic lupus erythematosus (sle): the skin rash, joint pain, and systemic manifestations may suggest lupus, particularly if accompanied by photosensitivity or other characteristic features.\\
3. psoriatic arthritis: if the rash resembles psoriasis, this type of arthritis could be considered, though it would more typically involve distinctive patterns of joint symptoms.}\\

\end{tiny}
\end{multicols}
\end{tcolorbox}

\begin{tcolorbox}[colback=gray!5!white,colframe=blue!50!white,title= Patient Diagnosis with Gemini 2.0 Flash \{LLM with RAG\} \{\textcolor{yellow}{Major Flaw}\}]
\begin{multicols}{2}
\begin{small}
\textbf{Patient Symptoms}\\

\textcolor{violet}{\textit{patient information:}}\\
- age: 61\\
- gender: female\\

\textcolor{violet}{\textit{problem description:}}\\
- primary problem: joint pain , joint swelling\\
- onset timing (in days): 0\\

\textcolor{violet}{\textit{symptoms and assessment:}}\\
- other symptoms: joint pain , joint swelling , morning stiffness\\
- fever history: yes, 100-102f ( , 100-102 f )\\
- joint pain: no\\
- swelling or deformity in joints: no\\
- redness in joints: not applicable\\
- warmth in joints: yes\\
- sleep disruption: yes\\
- hours of sleep: 5\\
- effect of physical activity on pain: increase\\
- effect of rest on pain: increase\\
- painkillers: no\\
- response to medication: not applicable\\
- skin rash: yes, hands / feet\\
- sunlight effect on rash: no\\
- grittiness in eyes: sand like sensation\\
- eye dryness (use of eye drops): yes\\
- difficulty swallowing dry foods: no\\
- difficulty sitting up: yes\\
- difficulty getting up from lying position: yes\\
- pain locations from image 1: right wrist, right mcp, right thigh, right knee, right ankle, right midfoot, unknown (×5)\\
- pain locations from image 2: upper back, right shoulder (post), middle back, lower midback\\
- prior diagnoses: yes, rheumatoid arthritis\\
- previous history of pain and swelling in wrists, small joints of hands and feet : yes\\
- current medications: yes, hypertension\\

\textcolor{violet}{\textit{doctor's diagnosis:}}\\
- final diagnosis: \textcolor{red}{\textbf{rheumatoid arthritis}}\\
\end{small}
\columnbreak

\textbf{Reasoning by Gemini 2.0 flash}\\
\begin{tiny}
\textbf{The highlighted passages have been identified by our clinical reviewers as instances of flawed or misleading reasoning in the model’s justification for an RA diagnosis.}\\
\textcolor{violet}{\textit{Reasons in support of rheumatoid arthritis diagnosis:}}\\
\texttt{prior diagnosis: the patient has a \hl{prior diagnosis of rheumatoid arthritis.}\\
joint pain and swelling: the patient presents with joint pain and swelling.\\
morning stiffness: the patient experiences morning stiffness.\\
joint involvement: pain locations include the right wrist, right mcp, right thigh, right knee, right ankle, and right midfoot. this involvement of the wrists and other peripheral joints supports the diagnosis. the patient also has a previous history of pain and swelling in wrists, small joints of hands and feet\\
warmth in joints: the patient reports warmth in the affected joints.\\
sleep disruption: the patient experiences sleep disruption, which can be associated with ra due to pain and discomfort.\\
eye dryness: the patient reports grittiness in eyes and uses eye drops, which is consistent with the potential association of ra with sjogren's syndrome (secondary dryness).\\
difficulty sitting/getting up: the patient experiences difficulty with physical activities like sitting up and getting up from a lying position, indicating functional limitations due to joint issues.\\
low grade fever: \hl{the patient has a fever (100-102f) which may be associated with rheumatoid arthritis.}}\\

\textcolor{violet}{\textit{Points against rheumatoid arthritis diagnosis:}}\\
\texttt{pain response to rest: \hl{pain increases with rest, which is atypical for inflammatory arthritis (typically improves with rest).}\\
skin rash: the patient has a skin rash on hands/feet, which is less common in ra and more suggestive of other connective tissue disorders.\\
absence of pain killer response: the patient hasn't taken any pain killers.\\
pain locations: the patient's pain locations include upper back, right shoulder (post), middle back, lower mid back which is against the diagnosis.\\
\hl{sunlight effect on rash: the patient doesn't have sunlight effect on rash which is against the diagnosis.}}\\

\textcolor{violet}{\textit{Explanation for considering rheumatoid arthritis:}}\\
\texttt{despite the points against, the patient's prior diagnosis, the pattern of joint involvement (wrists, mcps, knees, ankles), morning stiffness, and systemic symptoms like fatigue and eye dryness strongly suggest ra. the atypical pain response to rest and the presence of a rash warrant further investigation to rule out other conditions, but ra remains the most likely diagnosis given the overall clinical picture.}\\

\textcolor{violet}{\textit{Other possible differentials:}}\\
\texttt{psoriatic arthritis: this is a possibility due to the presence of joint pain and skin rash.\\
systemic lupus erythematosus (sle): joint pain, skin rash and fever can be seen in sle.\\
other connective tissue disorders: other mixed connective tissue disorders should also be considered.\\
sjögren's syndrome: this should be considered due to the presence of eye dryness.}\\

\end{tiny}
\end{multicols}
\end{tcolorbox}

\begin{tcolorbox}[colback=gray!5!white,colframe=blue!50!white,title= Patient Diagnosis with Gemini 2.0 Flash \{LLM with RAG\} \{\textcolor{yellow}{Minor Flaw}\}]
\begin{multicols}{2}
\begin{small}
\textbf{Patient Symptoms}\\

\textcolor{violet}{\textit{patient information:}}\\
- age: 39\\
- gender: female\\

\textcolor{violet}{\textit{problem description:}}\\
- primary problem: joint pain , joint swelling , inability to move hand / feet\\
- onset timing (in days): 0\\

\textcolor{violet}{\textit{symptoms and assessment:}}\\
- other symptoms: joint pain , joint swelling , inability to move hand/feet\\
- fever history: no\\
- joint pain: no , yes, lower backache\\
- swelling or deformity in joints: no\\
- redness in joints: no\\
- warmth in joints: no\\
- sleep disruption: yes\\
- hours of sleep: 6\\
- effect of physical activity on pain: increase\\
- effect of rest on pain: increase\\
- painkillers: yes\\
- response to medication: good relief\\
- skin rash: yes, hands / feet\\
- sunlight effect on rash: no\\
- grittiness in eyes: itching\\
- eye dryness (use of eye drops): no\\
- difficulty swallowing dry foods: no\\
- difficulty sitting up: yes\\
- difficulty getting up from lying position: yes\\
- pain locations from image 1: right shoulder, right mcp, right pip, right knee, left shoulder, unknown (×5)\\
- pain locations from image 2: left shoulder (post), right shoulder (post)\\
- prior diagnoses: others\\
- previous history of pain and swelling in wrists, small joints of hands and feet: no\\
- current medications: no\\

\textcolor{violet}{\textit{doctor's diagnosis:}}\\
- final diagnosis: \textcolor{red}{\textbf{Not rheumatoid arthritis}}\\
\end{small}
\columnbreak

\textbf{Reasoning by Gemini 2.0 flash}\\
\begin{tiny}
\textbf{The highlighted passages have been identified by our clinical reviewers as instances of flawed or misleading reasoning in the model’s justification for an RA diagnosis.}\\

\textcolor{violet}{\textit{Diagnosis: not rheumatoid arthritis}}\\

\textcolor{violet}{\textit{Reasons in support of this diagnosis:}}\\
\texttt{age discrepancy: the historical data describes patients aged 49, while the current patient is 39.\\
pain improvement with activity: the provided information does not contain this information\\
\hl{fever absence: the patient doesn't have a fever, while the historical data describes patients having fever.}\\
skin rash location: the patient has a skin rash on the hands and feet, while the historical data describes a skin rash on the face worsened by sunlight. this is not typical of rheumatoid arthritis.\\
grittiness in eyes: the patient has grittiness in eyes, while the historical data describes patients having no grittiness in eyes.\\
sleep disruption: \hl{the patient has sleep disruption, while the historical data describes patients having no sleep disruption.}\\
difficulty sitting up/getting up: the patient has difficulty with sitting up and getting up from a lying position, which isn't a prominent feature of rheumatoid arthritis.\\
pain locations: the patient has pain in both shoulders, while the historical data describes pain in the right mcp, right pip, right thigh, right knee, low back.\\
\hl{prior diagnoses: the patient does not have any prior diagnoses, while the historical data describes patients having rheumatoid arthritis.}\\
previous history: the patient does not have any previous history of pain and swelling in wrists, small joints of hands and feet, while the historical data describes patients having this history.}\\

\textcolor{violet}{\textit{Points against this diagnosis:}}\\
\texttt{joint pain and swelling: the patient presents with joint pain and swelling, which is a common symptom of rheumatoid arthritis.\\
good response to painkillers: the patient gets good relief with painkillers which is seen in rheumatoid arthritis}\\

\textcolor{violet}{\textit{Explanation for considering the diagnosis}}\\
\texttt{the combination of joint pain, swelling, skin rash, grittiness in the eyes, sleep disruption, and difficulty with movement suggests an alternative diagnosis. the absence of several key features from the historical data and the presence of other symptoms are more suggestive of a different connective tissue disorder or musculoskeletal condition.}\\

\textcolor{violet}{\textit{Other possible differentials:}}\\
\texttt{fibromyalgia: given the widespread pain, sleep disruption, and difficulty with movement, fibromyalgia is a strong consideration.\\
other connective tissue disorders (e.g., lupus): the skin rash, joint pain, and other symptoms could be indicative of another autoimmune connective tissue disease like lupus.}\\

\end{tiny}
\end{multicols}
\end{tcolorbox}

\begin{tcolorbox}[colback=gray!5!white,colframe=blue!50!white,title= Patient Diagnosis with GPT 4o \{LLM with RAG\} \{\textcolor{yellow}{Major Flaw}\}]
\begin{multicols}{2}
\begin{small}
\textbf{Patient Symptoms}\\

\textcolor{violet}{\textit{patient information:}}\\
- age: 45\\
- gender: female\\

\textcolor{violet}{\textit{problem description:}}\\
- primary problem: joint pain , joint swelling\\
- onset timing (in days): 0\\

\textcolor{violet}{\textit{symptoms and assessment:}}\\
- other symptoms: joint pain , joint swelling\\
- fever history: no\\
- joint pain: no\\
- swelling or deformity in joints: no\\
- redness in joints: no\\
- warmth in joints: no\\
- sleep disruption: no\\
- hours of sleep: 8\\
- effect of physical activity on pain: increase\\
- effect of rest on pain: reduce\\
- painkillers: no\\
- response to medication: not applicable\\
- skin rash: yes, hands / feet\\
- sunlight effect on rash: no\\
- grittiness in eyes: no\\
- eye dryness (use of eye drops): no\\
- difficulty swallowing dry foods: no\\
- difficulty sitting up: no\\
- difficulty getting up from lying position: no\\
- pain locations from image 1: right wrist, right mcp, right pip\\
- pain locations from image 2: no areas selected\\
- prior diagnoses: yes, rheumatoid arthritis\\
- previous history of pain and swelling in wrists, small joints of hands and feet: yes\\
- current medications: no\\

\textcolor{violet}{\textit{doctor's diagnosis:}}\\
- final diagnosis: \textcolor{red}{\textbf{rheumatoid arthritis}}\\
\end{small}
\columnbreak

\textbf{Reasoning by GPT 4o}\\
\begin{tiny}
\textbf{The highlighted passages have been identified by our clinical reviewers as instances of flawed or misleading reasoning in the model’s justification for an RA diagnosis.}\\

\textcolor{violet}{\textit{Reasons in support of rheumatoid arthritis diagnosis:}}\\
\texttt{1. involvement of joints: the patient has pain in the right wrist and small joints of the hands (mcp and pip joints), aligning with typical joint involvements seen in rheumatoid arthritis.\\
2. previous history: the patient has a previous history of pain and swelling in the wrists and small joints of the hands and feet, which is consistent with rheumatoid arthritis.\\
3. gender and age: the patient is a 45-year-old female, fitting the demographic commonly affected by rheumatoid arthritis.\\
4. \hl{impact of physical activity: pain increases with physical activity, indicating the possibility of inflammatory arthritis.}}\\

\textcolor{violet}{\textit{Points against rheumatoid arthritis diagnosis:}}\\
\texttt{1. skin rash: the presence of a skin rash on hands and feet is not characteristic of rheumatoid arthritis and suggests the possibility of another condition, such as a connective tissue disorder.\\
2. effect of rest on pain: the pain decreases with rest, which is atypical for inflammatory arthritis like rheumatoid arthritis, where pain usually improves with activity.\\
3. absence of morning stiffness: there's no mention of morning stiffness, a significant indicator of inflammatory arthritis.\\
4. no swelling or redness in joints: the patient does not present with swelling, warmth, or redness in the joints during assessment, which are common symptoms of rheumatoid arthritis.\\
5. no sleep disruption: unlike typical cases, the patient reports no sleep disruption and seems to have a full night's rest.}\\

\textcolor{violet}{\textit{Explanation for considering the diagnosis:}}\\
\texttt{the historical data with symptoms and the \hl{presence of previous diagnosis of rheumatoid arthritis} point toward this being the most likely diagnosis, considering the patient's profile and history of joint involvement. the patient's demographic and specific joint pain also support this diagnosis. however, the lack of certain symptoms like morning stiffness and joint swelling, and the presence of a skin rash, demand a thorough investigation and consideration of differential diagnoses.}\\

\textcolor{violet}{\textit{Other possible differential diagnoses:}}\\
\texttt{1. systemic lupus erythematosus (sle): given the presence of a skin rash and joint symptoms, sle could be a potential differential to consider.\\
2. psoriatic arthritis: if the skin rash is related to psoriasis, psoriatic arthritis should be considered.\\
3. palindromic rheumatism: this may account for intermittent joint pain and swelling episodes without continuous, chronic symptoms.\\
4. sjogren’s syndrome: though the patient does not present with dryness in eyes or mouth, sjogren’s can occasionally present atypically.\\
5. other connective tissue diseases: the presence of a skin rash and joint issues could also indicate other autoimmune conditions.}\\

\end{tiny}
\end{multicols}
\end{tcolorbox}

\begin{tcolorbox}[colback=gray!5!white,colframe=blue!50!white,title= Patient Diagnosis with GPT 4o \{LLM with RAG\} \{\textcolor{yellow}{Minor Flaw}\}]
\begin{multicols}{2}
\begin{small}
\textbf{Patient Symptoms}\\

\textcolor{violet}{\textit{patient information:}}\\
- age: 72\\
- gender: female\\

\textcolor{violet}{\textit{problem description:}}\\
- primary problem: joint pain , joint swelling , inability to move hand / feet\\
- onset timing (in days): 0\\

\textcolor{violet}{\textit{symptoms and assessment:}}\\
- other symptoms: joint pain , joint swelling , inability to move hand/feet, morning stiffness\\
- fever history: yes, less than 100f\\
- joint pain: no\\
- swelling or deformity in joints: no\\
- redness in joints: no\\
- warmth in joints: no\\
- sleep disruption: yes\\
- hours of sleep: 6\\
- effect of physical activity on pain: increase\\
- effect of rest on pain: increase\\
- painkillers: yes\\
- response to medication: moderate relief\\
- skin rash: yes, hands / feet\\
- sunlight effect on rash: no\\
- grittiness in eyes: itching\\
- eye dryness (use of eye drops): yes\\
- difficulty swallowing dry foods: no\\
- difficulty sitting up: yes\\
- difficulty getting up from lying position: yes\\
- pain locations from image 1: right shoulder, right upper arm, right wrist, right mcp, right pip, right knee, right ankle, right midfoot, left shoulder, left upper arm\\
- pain locations from image 2: posterior cervical region, left shoulder (post), lower back (left psis), left hip buttock, upper back, right shoulder (post), right psis\\
- prior diagnoses: yes, rheumatoid arthritis\\
- previous history of pain and swelling in wrists, small joints of hands and feet: yes\\
- current medications: yes, thyroid problem\\

\textcolor{violet}{\textit{doctor's diagnosis:}}\\
- final diagnosis: \textcolor{red}{\textbf{rheumatoid arthritis}}\\
\end{small}
\columnbreak

\textbf{Reasoning by GPT 4o}\\
\begin{tiny}
\textbf{The highlighted passages have been identified by our clinical reviewers as instances of flawed or misleading reasoning in the model’s justification for an RA diagnosis.}\\

\textcolor{violet}{\textit{Reasons supporting diagnosis of rheumatoid arthritis:}}\\
\texttt{1. presence of early morning stiffness: the patient reports morning stiffness, which is a common feature of inflammatory arthritis like rheumatoid arthritis.\\
2. involvement of wrists and small joints of hands or feet: previous history of pain and swelling in wrists and small joints is highly indicative of rheumatoid arthritis.\\
3. moderate relief with painkillers: the patient has moderate relief from medications, which is typical in rheumatoid arthritis management.\\
4. \hl{previous diagnoses: the patient has a prior diagnosis of rheumatoid arthritis}, adding credibility to this being a recurrent or chronic condition.\\
5. low-grade fever history: reports of fever below 100°f fit with low-grade fever occurrences sometimes seen in rheumatoid arthritis.\\
6. pain distribution: pain locations include many typical areas for rheumatoid arthritis involvement such as shoulders, arms, wrists, and the mcp and pip joints.}\\

\textcolor{violet}{\textit{Points against diagnosis of rheumatoid arthritis:}}\\
\texttt{1. pain worsens at rest: typically, rheumatoid arthritis pain improves with activity and worsens with rest, contrary to this patient's experience.\\
2. absence of warmth, swelling, or redness in joints: the current lack of swelling, warmth, and redness typically seen in active rheumatoid arthritis is against this diagnosis.\\
3. presence of skin rash: the presence of rash suggests other connective tissue disorders, as skin rashes are not characteristic of rheumatoid arthritis.\\
4. involvement of axial and mid-foot regions: pain is reported in the mid-foot and cervical region, which is atypical for early rheumatoid arthritis.\\
5. dryness and eye symptoms: eye symptoms might suggest another overlapping autoimmune disorder such as sjogren’s syndrome.}\\

\textcolor{violet}{\textit{Explanation for considering the diagnosis:}}\\
\texttt{despite the current atypical aspects against rheumatoid arthritis, \hl{the history of previous effective diagnosis} and treatment, along with characteristic symptoms like morning stiffness, joint pain, and prior specific joint involvement, strongly suggests an underlying ongoing rheumatoid arthritis condition. the presence of previous symptoms aligning with rheumatoid arthritis and response to medications further strengthen the probability of this diagnosis.}\\

\textcolor{violet}{\textit{Other possible differentials:}}\\
\texttt{1. systemic lupus erythematosus (sle): could explain the skin rash and diffuse joint pain, along with the eye symptoms.\\
2. sjogren's syndrome: the dryness of eyes and other autoimmune symptoms suggest possible overlap or concurrent sjogren’s syndrome.\\
3. psoriatic arthritis: given the skin rash and joint involvement, psoriatic arthritis could be considered, though lacks classical psoriasis lesions.}\\

\end{tiny}
\end{multicols}
\end{tcolorbox}

\begin{tcolorbox}[colback=gray!5!white,colframe=blue!50!white,title= Patient Diagnosis with Gemini 2.0 Flash \{two agents with RAG\} \{\textcolor{yellow}{Major Flaw}\}]
\begin{multicols}{2}
\begin{small}
\textbf{Patient Symptoms}\\

\textcolor{violet}{\textit{patient information:}}\\
- age: 45\\
- gender: female\\

\textcolor{violet}{\textit{problem description:}}\\
- primary problem: joint pain , joint swelling\\
- onset timing (in days): 0\\

\textcolor{violet}{\textit{symptoms and assessment:}}\\
- other symptoms: joint pain , joint swelling\\
- fever history: no\\
- joint pain: no\\
- swelling or deformity in joints: no\\
- redness in joints: no\\
- warmth in joints: no\\
- sleep disruption: no\\
- hours of sleep: 8\\
- effect of physical activity on pain: increase\\
- effect of rest on pain: reduce\\
- painkillers: no\\
- response to medication: not applicable\\
- skin rash: yes, hands / feet\\
- sunlight effect on rash: no\\
- grittiness in eyes: no\\
- eye dryness (use of eye drops): no\\
- difficulty swallowing dry foods: no\\
- difficulty sitting up: no\\
- difficulty getting up from lying position: no\\
- pain locations from image 1: right wrist, right mcp, right pip\\
- pain locations from image 2: no areas selected\\
- prior diagnoses: yes, rheumatoid arthritis\\
- previous history of pain and swelling in wrists, small joints of hands and feet: yes\\
- current medications: no\\

\textcolor{violet}{\textit{doctor's diagnosis:}}\\
- final diagnosis: \textcolor{red}{\textbf{rheumatoid arthritis}}\\
\end{small}
\columnbreak

\textbf{Reasoning by Gemini 2.0 flash}\\
\begin{tiny}
\textbf{The highlighted passages have been identified by our clinical reviewers as instances of flawed or misleading reasoning in the model’s justification for an RA diagnosis.}\\

\textcolor{violet}{\textit{Reasons in support of rheumatoid arthritis diagnosis:}}\\
\texttt{1. joint pain and swelling: the patient presents with joint pain and swelling, which are primary symptoms of rheumatoid arthritis.\\
2. pain locations: specific pain locations in the right wrist, right mcp, and right pip joints are consistent with ra.\\
3. effect of physical activity/rest: \hl{increased pain with physical activity and reduced pain with rest align with ra symptoms.}\\
4. prior diagnoses: \hl{the patient has a prior diagnosis of rheumatoid arthritis.}\\
5. previous history: a history of pain and swelling in wrists and small joints of hands and feet supports the diagnosis.}\\

\textcolor{violet}{\textit{Points against rheumatoid arthritis diagnosis:}}\\
\texttt{1. absence of fever: the patient reports no fever history, although low-grade fever can be associated with ra.\\
2. absence of joint swelling/deformity/redness/warmth: the patient denies swelling or deformity in joints, redness in joints, and warmth in joints. these are common signs of ra.\\
3. absence of sleep disruption: the patient reports 8 hours of sleep, indicating no sleep disruption, \hl{which is contrary to the historical data.}\\
4. skin rash: the presence of a skin rash on hands/feet is not typical for ra and suggests other connective tissue disorders.\\
5. painkillers: the patient is not taking painkillers.\\
6. response to medication: response to medication is not applicable.\\
7. difficulty sitting up: the patient denies difficulty sitting up.}\\

\textcolor{violet}{\textit{Explanation for considering the diagnosis:}}\\
\texttt{despite some contradictory symptoms, the patient's prior diagnosis of rheumatoid arthritis, along with the presence of joint pain and swelling in specific locations (right wrist, right mcp, right pip), and the effect of physical activity and rest on pain, suggests that rheumatoid arthritis should be considered. the previous history of pain and swelling in wrists, and small joints of hands and feet further supports this.}\\

\textcolor{violet}{\textit{Other possible differentials:}}\\
\texttt{1. other connective tissue disorders: the presence of a skin rash suggests other connective tissue disorders such as: systemic lupus erythematosus (sle), psoriatic arthritis.\\
2. osteoarthritis: while less likely given the specific joint involvement, osteoarthritis could be considered if the inflammatory markers are negative.}\\

\end{tiny}
\end{multicols}
\end{tcolorbox}

\begin{tcolorbox}[colback=gray!5!white,colframe=blue!50!white,title= Patient Diagnosis with Gemini 2.0 Flash \{two agents with RAG\} \{\textcolor{yellow}{Minor Flaw}\}]
\begin{multicols}{2}
\begin{small}
\textbf{Patient Symptoms}\\

\textcolor{violet}{\textit{patient information:}}\\
- age: 51\\
- gender: female\\

\textcolor{violet}{\textit{problem description:}}\\
- primary problem: joint pain , joint swelling , inability to move hand / feet , morning stiffness\\
- onset timing (in days): 0\\

\textcolor{violet}{\textit{symptoms and assessment:}}\\
- other symptoms: joint pain , joint swelling , inability to move hand/feet , morning stiffness\\
- fever history: yes, less than 100f\\
- joint pain: others\\
- swelling or deformity in joints: yes, left hand, right hand\\
- redness in joints: yes\\
- warmth in joints: yes\\
- sleep disruption: no\\
- hours of sleep: 10\\
- effect of physical activity on pain: increase\\
- effect of rest on pain: increase\\
- painkillers: yes\\
- response to medication: mild relief\\
- skin rash: yes, hands / feet\\
- sunlight effect on rash: no\\
- grittiness in eyes: sand like sensation\\
- eye dryness (use of eye drops): no\\
- difficulty swallowing dry foods: no\\
- difficulty sitting up: yes\\
- difficulty getting up from lying position: yes\\
- pain locations from image 1: right shoulder, right wrist, right mcp, right pip, right knee, right ankle, right midfoot, left shoulder\\
- pain locations from image 2: left shoulder (post), left elbow, left heel, upper back, right shoulder (post), right elbow\\
- prior diagnoses: yes, rheumatoid arthritis\\
- previous history of pain and swelling in wrists, small joints of hands and feet: yes\\
- current medications: no\\

\textcolor{violet}{\textit{doctor's diagnosis:}}\\
- final diagnosis: \textcolor{red}{\textbf{rheumatoid arthritis}}\\
\end{small}
\columnbreak

\textbf{Reasoning by Gemini 2.0 flash}\\
\begin{tiny}
\textbf{The highlighted passages have been identified by our clinical reviewers as instances of flawed or misleading reasoning in the model’s justification for an RA diagnosis.}\\

\textcolor{violet}{\textit{Reasons in support of rheumatoid arthritis diagnosis:}}\\
\texttt{1. age and gender: the patient is a 51-year-old female, which aligns with the typical demographic for rheumatoid arthritis onset.\\
2. joint pain and swelling: the patient presents with joint pain, swelling, and inability to move hands/feet, which are hallmark symptoms of rheumatoid arthritis.\\
3. morning stiffness: the presence of morning stiffness, a key indicator of inflammatory arthritis, supports the diagnosis.\\
4. effect of activity and rest on pain: \hl{pain increases with both physical activity and rest, which is more indicative of inflammatory arthritis.}\\
5. joint involvement: the patient experiences pain in multiple joints, including the wrists, mcps, pips, shoulders, knees, and ankles. this pattern of peripheral joint involvement is typical of rheumatoid arthritis.\\
6. previous history: the patient has a previous history of pain and swelling in the wrists and small joints of the hands and feet, which aligns with rheumatoid arthritis.\\
7. prior diagnosis: \hl{the patient has a prior diagnosis of rheumatoid arthritis.}\\
8. warmth and redness in joints: presence of warmth and redness in the affected joints indicates inflammation.}\\

\textcolor{violet}{\textit{Points against rheumatoid arthritis diagnosis:}}\\
\texttt{1. sleep disruption: the patient reports no sleep disruption, which is atypical as pain often leads to disturbed sleep in rheumatoid arthritis patients.\\
2. painkiller response: the patient experiences only mild relief from painkillers, which is less typical of rheumatoid arthritis, where nsaids often provide moderate relief.\\
3. skin rash: the patient has skin rash on hands and feet, which is not typically associated with rheumatoid arthritis.\\
4. fever: the patient presents with low grade fever, which may be associated with rheumatoid arthritis, but high-grade fever is unlikely.\\
5. grittiness in eyes: the patient reports grittiness in eyes, which is not typically associated with rheumatoid arthritis.\\
6. hours of sleep: the patient sleeps for 10 hours, which is not typical as pain often leads to disturbed sleep in rheumatoid arthritis patients.}\\

\textcolor{violet}{\textit{Explanation for considering the diagnosis:}}\\
\texttt{despite some atypical features like the skin rash and mild painkiller response, the overall clinical picture strongly suggests rheumatoid arthritis. the patient's age, gender, joint pain, swelling, morning stiffness, pattern of joint involvement, and prior history are all consistent with the disease. the grittiness in eyes can be associated with rheumatoid arthritis. the low grade fever may be associated with rheumatoid arthritis. the skin rash can be a secondary symptom due to rheumatoid arthritis. the effect of rest and activity on pain can be associated with rheumatoid arthritis.}\\

\textcolor{violet}{\textit{Other possible differentials:}}\\
\texttt{1. systemic lupus erythematosus (sle): this autoimmune disease can present with joint pain, skin rash, fatigue, and other systemic symptoms.\\
2. psoriatic arthritis: this inflammatory arthritis is associated with psoriasis, a skin condition characterized by scaly, red patches.\\
3. other connective tissue disorders: other conditions like mixed connective tissue disease (mctd) or sjogren's syndrome could also be considered, especially given the eye symptoms.}\\

\end{tiny}
\end{multicols}
\end{tcolorbox}

\begin{tcolorbox}[colback=gray!5!white,colframe=blue!50!white,title= Patient Diagnosis with GPT 4o \{Two agents with RAG\} \{\textcolor{yellow}{Major Flaw}\}]
\begin{multicols}{2}
\begin{small}
\textbf{Patient Symptoms}\\

\textcolor{violet}{\textit{patient information:}}\\
- age: 45\\
- gender: female\\

\textcolor{violet}{\textit{problem description:}}\\
- primary problem: joint pain , joint swelling\\
- onset timing (in days): 0\\

\textcolor{violet}{\textit{symptoms and assessment:}}\\
- other symptoms: joint pain , joint swelling\\
- fever history: no\\
- joint pain: no\\
- swelling or deformity in joints: no\\
- redness in joints: no\\
- warmth in joints: no\\
- sleep disruption: no\\
- hours of sleep: 8\\
- effect of physical activity on pain: increase\\
- effect of rest on pain: reduce\\
- painkillers: no\\
- response to medication: not applicable\\
- skin rash: yes, hands / feet\\
- sunlight effect on rash: no\\
- grittiness in eyes: no\\
- eye dryness (use of eye drops): no\\
- difficulty swallowing dry foods: no\\
- difficulty sitting up: no\\
- difficulty getting up from lying position: no\\
- pain locations from image 1: right wrist, right mcp, right pip\\
- pain locations from image 2: no areas selected\\
- prior diagnoses: yes, rheumatoid arthritis\\
- previous history of pain and swelling in wrists, small joints of hands and feet: yes\\
- current medications: no\\

\textcolor{violet}{\textit{doctor's diagnosis:}}\\
- final diagnosis: \textcolor{red}{\textbf{rheumatoid arthritis}}\\
\end{small}
\columnbreak

\textbf{Reasoning by GPT 4o}\\
\begin{tiny}
\textbf{The highlighted passages have been identified by our clinical reviewers as instances of flawed or misleading reasoning in the model’s justification for an RA diagnosis.}\\

\textcolor{violet}{\textit{Reasons in support of the diagnosis of rheumatoid arthritis:}}\\
\texttt{1. involvement of specific joints: the patient has pain and swelling in joints like the right wrist, right mcp, and right pip, which aligns with rheumatoid arthritis, as it often affects wrists and small joints of the hands.\\
2. previous history: the patient has a prior history of pain and swelling in wrists and small joints of hands and feet, which is consistent with rheumatoid arthritis.\\
3. onset and progression: \hl{the pain increases with activity and is reduced by rest. though atypical, it can show variability in pain patterns.}\\
4. lack of deformity evolution: gradual evolution of deformities is common in rheumatoid arthritis, though not explicitly stated, prior diagnosis and joint involvement suggest ongoing progression.\\
5. effect on pain with rest: \hl{the pain reduces with rest, which aligns with the typical pattern for inflammatory arthritis, including ra.}}\\

\textcolor{violet}{\textit{Points against the diagnosis of rheumatoid arthritis:}}\\
\texttt{1. skin rash presence: the patient reports a skin rash on hands/feet, which is not typical for rheumatoid arthritis and could suggest a different connective tissue disorder.\\
2. absence of fever and response to painkillers: the patient reports no history of fever and no current medication use, indicating a lack of pain management typically associated with ra.\\
3. no eye or mouth dryness: dryness of eyes or mouth is not observed, which, while not necessary, can be associated with ra and sjögren's syndrome.}\\

\textcolor{violet}{\textit{Explanation for considering the diagnosis:}}\\
\texttt{the diagnosis of rheumatoid arthritis is considered due to the specific joint involvement and history of symptoms, which is consistent with the disease. the patient’s symptom history, including the presence of pain in specific joints with a pattern of activity-related increase and relief with rest, further supports the diagnosis. previous diagnoses and symptom continuation align with rheumatoid arthritis manifestations, despite some atypical features like skin rash.}\\

\textcolor{violet}{\textit{Other possible differential diagnosis:}}\\
\texttt{1. systemic lupus erythematosus (sle): the presence of a skin rash and joint symptoms could suggest sle, which can present with similar musculoskeletal symptoms and skin manifestations.\\
2. psoriatic arthritis: considered if there is a family history or signs of psoriasis, which can cause joint inflammation and skin involvement.\\
3. mixed connective tissue disease (mctd): the presence of rash alongside joint symptoms might suggest mctd, encompassing features of multiple connective tissue disorders.}\\

\end{tiny}
\end{multicols}
\end{tcolorbox}

\begin{tcolorbox}[colback=gray!5!white,colframe=blue!50!white,title= Patient Diagnosis with GPT 4o \{Two agents with RAG\} \{\textcolor{yellow}{Minor Flaw}\}]
\begin{multicols}{2}
\begin{small}
\textbf{Patient Symptoms}\\

\textcolor{violet}{\textit{patient information:}}\\
- age: 31\\
- gender: female\\

\textcolor{violet}{\textit{problem description:}}\\
- primary problem: joint pain\\
- onset timing (in days): 0\\

\textcolor{violet}{\textit{symptoms and assessment:}}\\
- other symptoms: joint pain\\
- fever history: yes, less than 100f\\
- joint pain: no\\
- swelling or deformity in joints: no\\
- redness in joints: yes\\
- warmth in joints: no\\
- sleep disruption: no\\
- hours of sleep: 8\\
- effect of physical activity on pain: increase\\
- effect of rest on pain: reduce\\
- painkillers: no\\
- response to medication: not applicable\\
- skin rash: no\\
- sunlight effect on rash: no\\
- grittiness in eyes: no\\
- eye dryness (use of eye drops): no\\
- difficulty swallowing dry foods: no\\
- difficulty sitting up: no\\
- difficulty getting up from lying position: no\\
- pain locations from image 1: right mcp, right pip, right knee, unknown\\
- pain locations from image 2: left elbow, right elbow\\
- prior diagnoses: others\\
- previous history of pain and swelling in wrists, small joints of hands and feet: yes\\
- current medications: no\\

\textcolor{violet}{\textit{doctor's diagnosis:}}\\
- final diagnosis: \textcolor{red}{\textbf{Not rheumatoid arthritis}}\\
\end{small}
\columnbreak

\textbf{Reasoning by GPT 4o}\\
\begin{tiny}
\textbf{The highlighted passages have been identified by our clinical reviewers as instances of flawed or misleading reasoning in the model’s justification for an RA diagnosis.}\\

\textcolor{violet}{\textit{Reasons in support of rheumatoid arthritis diagnosis:}}\\
\texttt{1. previous history: the patient has a previous history of pain and swelling in wrists and small joints of hands, which is consistent with rheumatoid arthritis.\\
2. joint involvement: involvement of small joints such as the right mcp and pip is noted. this is characteristic of rheumatoid arthritis.\\
3. pain pattern: pain increases with physical activity and reduces with rest, which is a common pattern in rheumatoid arthritis.\\
4. redness in joints: presence of redness in joints is a symptom that can be associated with inflammation characteristic of rheumatoid arthritis, although not definitive on its own.\\
5. low-grade fever: the presence of low-grade fever is consistent with the potential systemic inflammatory response seen in rheumatoid arthritis.}\\

\textcolor{violet}{\textit{Points against rheumatoid arthritis diagnosis:}}\\
\texttt{1. lack of joint swelling: the absence of swelling or deformity in joints contradicts a typical rheumatoid arthritis presentation where swelling is often prominent.\\
2. no painkillers: the patient is not on painkillers, so we do not have information on their response to such medication, which typically provides relief in rheumatoid arthritis.\\
3. diagnosis by doctor: \hl{the final diagnosis given by the doctor is fibromyalgia, not rheumatoid arthritis.}\\
4. effect of activity and rest: typically in rheumatoid arthritis, pain increases after rest and diminishes with activity. the patient reports the opposite.\\
5. fibromyalgia indication: fibromyalgia often presents with widespread pain and can correlate with symptoms this patient is experiencing, distinct from the localized joint issues of rheumatoid arthritis.}\\

\textcolor{violet}{\textit{Explanation for considering the diagnosis:}}\\
\texttt{the diagnosis of rheumatoid arthritis was considered primarily due to the previous history of joint pain and swelling, the specific joints involved (e.g., mcp and pip), and some systemic symptoms like low-grade fever. however, the lack of swelling, response to treatment, and doctor’s final diagnosis suggest that rheumatoid arthritis may not be the best fit.}\\

\textcolor{violet}{\textit{Other possible differential diagnoses:}}\\
\texttt{1. fibromyalgia: given the doctor’s diagnosis, fibromyalgia is characterized by widespread musculoskeletal pain, which might align more closely with the patient's symptoms, especially with the absence of swelling and redness.\\
2. connective tissue disorder: other connective tissue disorders could present similarly but would typically include additional symptoms not observed here.\\
3. psoriatic arthritis: involvement of multiple joints and some of the symptoms could align with psoriatic arthritis, but the absence of skin changes or typical nail involvement may rule it out.\\
4. osteoarthritis: while more common in older adults, involvement of smaller joints can occur, yet it usually presents with different pattern of symptoms.}\\

\end{tiny}
\end{multicols}
\end{tcolorbox}

\begin{tcolorbox}[colback=gray!5!white,colframe=blue!50!white,title= Patient Diagnosis with Gemini 2.0 Flash \{Three agents with RAG\} \{\textcolor{yellow}{Major Flaw}\}]
\begin{multicols}{2}
\begin{small}
\textbf{Patient Symptoms}\\

\textcolor{violet}{\textit{patient information:}}\\
- age: 45\\
- gender: female\\

\textcolor{violet}{\textit{problem description:}}\\
- primary problem: joint pain, joint swelling\\
- onset timing (in days): 0\\

\textcolor{violet}{\textit{symptoms and assessment:}}\\
- other symptoms: joint pain, joint swelling\\
- fever history: no\\
- joint pain: no\\
- swelling or deformity in joints: no\\
- redness in joints: no\\
- warmth in joints: no\\
- sleep disruption: no\\
- hours of sleep: 8\\
- effect of physical activity on pain: increase\\
- effect of rest on pain: reduce\\
- painkillers: no\\
- response to medication: not applicable\\
- skin rash: yes, hands / feet\\
- sunlight effect on rash: no\\
- grittiness in eyes: no\\
- eye dryness (use of eye drops): no\\
- difficulty swallowing dry foods: no\\
- difficulty sitting up: no\\
- difficulty getting up from lying position: no\\
- pain locations from image 1: right wrist, right mcp, right pip\\
- pain locations from image 2: no areas selected\\
- prior diagnoses: yes, rheumatoid arthritis\\
- previous history of pain and swelling in wrists, small joints of hands and feet: yes\\
- current medications: no\\

\textcolor{violet}{\textit{doctor's diagnosis:}}\\
- final diagnosis: \textcolor{red}{\textbf{rheumatoid arthritis}}\\
\end{small}
\columnbreak

\textbf{Reasoning by Gemini 2.0 flash}\\
\begin{tiny}
\textbf{The highlighted passages have been identified by our clinical reviewers as instances of flawed or misleading reasoning in the model’s justification for an RA diagnosis.}\\

\textcolor{violet}{\textit{Reasons in support of rheumatoid arthritis diagnosis:}}\\
\texttt{joint pain and swelling: the patient presents with joint pain and swelling, the primary symptoms of rheumatoid arthritis.\\
\hl{pain increased with physical activity, reduced with rest: this pattern aligns with the inflammatory nature of ra}.\\
pain locations: the pain locations (right wrist, right mcp, right pip) are typical of ra, which commonly affects small joints of the hands and wrists.\\
previous history: the patient has a previous history of pain and swelling in wrists and small joints of hands and feet, strongly suggestive of ra.\\
\hl{prior diagnosis: patient has prior diagnosis of rheumatoid arthritis}.}\\

\textcolor{violet}{\textit{Points against rheumatoid arthritis diagnosis:}}\\
\texttt{skin rash: the presence of a skin rash on hands/feet is atypical for ra and raises suspicion for other connective tissue disorders.\\
fever history: no fever history is present in the patient, but low-grade fever may be associated with rheumatoid arthritis.}\\

\textcolor{violet}{\textit{Explanation for considering the diagnosis:}}\\
\texttt{despite the presence of a skin rash, the constellation of symptoms, including joint pain, swelling, \hl{pain patterns}, affected joint locations, and previous history, strongly points towards ra. the skin rash needs further investigation, but it doesn't necessarily exclude ra.}\\

\textcolor{violet}{\textit{Other possible differentials:}}\\
\texttt{systemic lupus erythematosus (sle): given the skin rash, sle is a strong differential. sle can present with joint pain, swelling, and skin manifestations.\\
psoriatic arthritis: this should be considered if there are any signs of psoriasis (skin or nail changes).\\
other connective tissue diseases: other less common connective tissue diseases should also be considered, especially if further investigations reveal additional systemic features.}\\

\end{tiny}
\end{multicols}
\end{tcolorbox}

\begin{tcolorbox}[colback=gray!5!white,colframe=blue!50!white,title= Patient Diagnosis with Gemini 2.0 Flash \{LLM with no RAG\} \{\textcolor{yellow}{Minor Flaw}\}]
\begin{multicols}{2}
\begin{small}
\textbf{Patient Symptoms}\\

\textcolor{violet}{\textit{patient information:}}\\
- age: 72\\
- gender: female\\

\textcolor{violet}{\textit{problem description:}}\\
- primary problem: joint pain, joint swelling, inability to move hand/feet\\
- onset timing (in days): 0\\

\textcolor{violet}{\textit{symptoms and assessment:}}\\
- other symptoms: joint pain, joint swelling, inability to move hand/feet, morning stiffness\\
- fever history: yes, less than 100f\\
- joint pain: no\\
- swelling or deformity in joints: no\\
- redness in joints: no\\
- warmth in joints: no\\
- sleep disruption: yes\\
- hours of sleep: 6\\
- effect of physical activity on pain: increase\\
- effect of rest on pain: increase\\
- painkillers: yes\\
- response to medication: moderate relief\\
- skin rash: yes, hands / feet\\
- sunlight effect on rash: no\\
- grittiness in eyes: itching\\
- eye dryness (use of eye drops): yes\\
- difficulty swallowing dry foods: no\\
- difficulty sitting up: yes\\
- difficulty getting up from lying position: yes\\
- pain locations from image 1: right shoulder, right upper arm, right wrist, right mcp, right pip, right knee, right ankle, right midfoot, left shoulder, left upper arm\\
- pain locations from image 2: posterior cervical region, left shoulder (post), lower back (left psis), left hip buttock, upper back, right shoulder (post), right psis\\
- prior diagnoses: yes, rheumatoid arthritis\\
- previous history of pain and swelling in wrists, small joints of hands and feet: yes\\
- current medications: yes, thyroid problem\\

\textcolor{violet}{\textit{doctor's diagnosis:}}\\
- final diagnosis: \textcolor{red}{\textbf{rheumatoid arthritis}}\\
\end{small}
\columnbreak

\textbf{Reasoning by Gemini 2.0 flash}\\
\begin{tiny}
\textbf{The highlighted passages have been identified by our clinical reviewers as instances of flawed or misleading reasoning in the model’s justification for an RA diagnosis.}\\

\textcolor{violet}{\textit{Reasons in support of rheumatoid arthritis diagnosis:}}\\
\texttt{joint involvement: the patient presents with joint pain, swelling, and inability to move hands/feet, which are primary symptoms of ra. specific joints mentioned include right shoulder, right upper arm, right wrist, right mcp, right pip, right knee, right ankle, right midfoot, left shoulder.\\
morning stiffness: the patient experiences morning stiffness, a key characteristic of inflammatory arthritis like ra.\\
sleep disruption: sleep disruption is common in ra due to pain and discomfort.\\
previous history: \hl{the patient has a prior diagnosis of rheumatoid arthritis} and a history of pain and swelling in wrists, small joints of hands and feet, strongly suggesting a recurrence or continuation of the condition.\\
eye dryness: the patient has eye dryness and use of eye drops, which can be associated with rheumatoid arthritis.\\
difficulty: difficulty sitting up and getting up from lying position is reported.}\\

\textcolor{violet}{\textit{Points against rheumatoid arthritis diagnosis:}}\\
\texttt{joint pain assessment: the patient reports no joint pain or swelling/deformity.\\
effect of physical activity and rest on pain: pain increases with both physical activity and rest, which is not typical for \hl{ra where pain usually improves with rest.}\\
painkiller response: the patient's response to medication is only moderate relief, while ra patients often experience better pain relief from painkillers.\\
skin rash: the patient has a skin rash on hands/feet, which is not typical of ra and could suggest another connective tissue disorder.\\
absence of joint redness and warmth: the patient reports no redness or warmth in the joints.}\\

\textcolor{violet}{\textit{Explanation for considering the diagnosis:}}\\
\texttt{despite some contradictory symptoms, the prior diagnosis of ra, the presence of morning stiffness, joint pain, joint swelling, inability to move hand/feet, sleep disruption, eye dryness, difficulty sitting up and getting up from lying position and previous history of pain and swelling in wrists, small joints of hands and feet is a strong indicator. the skin rash, lack of joint redness and warmth, and unusual pain response to activity/rest warrant further investigation. \hl{the "no" joint pain and swelling/deformity need to be clarified with the patient. the current symptoms could be an ra flare with atypical features or a co-existing condition.}}\\

\textcolor{violet}{\textit{Other possible differentials:}}\\
\texttt{other connective tissue disorders: given the skin rash, conditions like lupus or psoriatic arthritis should be considered.\\
osteoarthritis: although less likely given the inflammatory symptoms, oa could be a contributing factor, especially with the patient's age.\\
fibromyalgia: this could explain widespread pain, sleep disruption, and unusual pain response to activity/rest.\\
sjögren's syndrome: this could explain the eye dryness and joint pain.}\\

\end{tiny}
\end{multicols}
\end{tcolorbox}

\begin{tcolorbox}[colback=gray!5!white,colframe=blue!50!white,title= Patient Diagnosis with GPT 4o \{Three agents with RAG\} \{\textcolor{yellow}{Major Flaw}\}]
\begin{multicols}{2}
\begin{small}
\textbf{Patient Symptoms}\\

\textcolor{violet}{\textit{patient information:}}\\
- age: 45\\
- gender: female\\

\textcolor{violet}{\textit{problem description:}}\\
- primary problem: joint pain, joint swelling\\
- onset timing (in days): 0\\

\textcolor{violet}{\textit{symptoms and assessment:}}\\
- other symptoms: joint pain, joint swelling\\
- fever history: no\\
- joint pain: no\\
- swelling or deformity in joints: no\\
- redness in joints: no\\
- warmth in joints: no\\
- sleep disruption: no\\
- hours of sleep: 8\\
- effect of physical activity on pain: increase\\
- effect of rest on pain: reduce\\
- painkillers: no\\
- response to medication: not applicable\\
- skin rash: yes, hands / feet\\
- sunlight effect on rash: no\\
- grittiness in eyes: no\\
- eye dryness (use of eye drops): no\\
- difficulty swallowing dry foods: no\\
- difficulty sitting up: no\\
- difficulty getting up from lying position: no\\
- pain locations from image 1: right wrist, right mcp, right pip\\
- pain locations from image 2: no areas selected\\
- prior diagnoses: yes, rheumatoid arthritis\\
- previous history of pain and swelling in wrists, small joints of hands and feet: yes\\
- current medications: no\\

\textcolor{violet}{\textit{doctor's diagnosis:}}\\
- final diagnosis: \textcolor{red}{\textbf{rheumatoid arthritis}}\\
\end{small}
\columnbreak

\textbf{Reasoning by GPT 4o}\\
\begin{tiny}
\textbf{The highlighted passages have been identified by our clinical reviewers as instances of flawed or misleading reasoning in the model’s justification for an RA diagnosis.}\\

\textcolor{violet}{\textit{Reasons in support of this diagnosis:}}\\
\texttt{1. joint pain and swelling, particularly in the wrist and small joints of the hands, which are classically involved in rheumatoid arthritis.\\
2. \hl{prior diagnosis of rheumatoid arthritis} and previous history of pain and swelling in the wrists, small joints of hands, and feet.\\
3. pain locations specified in the assessment align with common areas affected by rheumatoid arthritis, namely the right wrist, right mcp, and right pip joints.\\
4. \hl{pain increases with physical activity but is reduced by rest, which aligns with inflammatory joint disease.}\\
5. absence of axial (spine) involvement, in line with rheumatoid arthritis patterns.}\\

\textcolor{violet}{\textit{Points against this diagnosis:}}\\
\texttt{1. presence of a skin rash, particularly on the hands and feet, which is not typical for rheumatoid arthritis and might suggest other connective tissue disorders.\\
2. no mention of early morning stiffness, which is a classic symptom of rheumatoid arthritis.\\
3. no reported warmth in the joints during the assessment phase, which is commonly present in active rheumatoid arthritis.\\
4. joint pain reportedly does not exist currently according to the problem description, which contrasts with the final diagnosis.}\\

\textcolor{violet}{\textit{Explanation for considering the diagnosis:}}\\
\texttt{the diagnosis of rheumatoid arthritis is supported by the history of joint pain and swelling in typical rheumatoid arthritis-affected areas, coupled with a previous diagnosis of the condition. despite the lack of early morning stiffness and the presence of a skin rash, the involvement of specific joints and the history of similar symptoms point towards rheumatoid arthritis. the problems with redness, warmth, and current pain need consideration as potential indicators of disease severity or remission phase.}\\

\textcolor{violet}{\textit{Other potential differentials for the patient:}}\\
\texttt{1. psoriatic arthritis: given the presence of a skin rash on hands and feet, psoriatic arthritis can be considered. it often presents with joint pain and rash.\\
2. systemic lupus erythematosus (sle): characterized by joint pain and skin rashes, particularly if sun exposure does not affect the rash.\\
3. sjögren's syndrome: if dryness symptoms were present, though they are not reported in this case, along with arthritis, it would be considered.}\\

\end{tiny}
\end{multicols}
\end{tcolorbox}

\begin{tcolorbox}[colback=gray!5!white,colframe=blue!50!white,title= Patient Diagnosis with GPT 4o \{Three agents with RAG\} \{\textcolor{yellow}{Minor Flaw}\}]
\begin{multicols}{2}
\begin{small}
\textbf{Patient Symptoms}\\

\textcolor{violet}{\textit{patient information:}}\\
- age: 48\\
- gender: male\\

\textcolor{violet}{\textit{problem description:}}\\
- primary problem: joint pain, joint swelling, morning stiffness\\
- onset timing (in days): 0\\

\textcolor{violet}{\textit{symptoms and assessment:}}\\
- other symptoms: joint pain, joint swelling, inability to move hand/feet, morning stiffness\\
- fever history: yes, less than 100f\\
- joint pain: yes, lower backache\\
- swelling or deformity in joints: no\\
- redness in joints: no\\
- warmth in joints: yes\\
- sleep disruption: yes\\
- hours of sleep: 5\\
- effect of physical activity on pain: increase\\
- effect of rest on pain: increase\\
- painkillers: yes\\
- response to medication: mild relief\\
- skin rash: no\\
- sunlight effect on rash: no\\
- grittiness in eyes: no\\
- eye dryness (use of eye drops): no\\
- difficulty swallowing dry foods: no\\
- difficulty sitting up: yes\\
- difficulty getting up from lying position: yes\\
- pain locations from image 1: right shoulder, right wrist, right mcp\\
- pain locations from image 2: left elbow, lower back (left PSIS), upper back, right shoulder (post), right elbow, right PSIS\\
- prior diagnoses: yes, rheumatoid arthritis\\
- previous history of pain and swelling in wrists, small joints of hands and feet: yes\\
- current medications: yes, hypertension\\

\textcolor{violet}{\textit{doctor's diagnosis:}}\\
- final diagnosis: \textcolor{red}{\textbf{rheumatoid arthritis}}\\
\end{small}
\columnbreak

\textbf{Reasoning by GPT 4o}\\
\begin{tiny}
\textbf{The highlighted passages have been identified by our clinical reviewers as instances of flawed or misleading reasoning in the model’s justification for an RA diagnosis.}\\

\textcolor{violet}{\textit{Reasons in support of this diagnosis:}}\\
\texttt{1. presence of early morning stiffness: the patient reports morning stiffness, which is a characteristic symptom of inflammatory arthritis like rheumatoid arthritis.\\
2. pain worsens after rest: the pain and stiffness worsening with rest aligns with rheumatoid arthritis, although in this case, activity also increases pain which is somewhat atypical.\\
3. involvement of joints: the patient experiences joint pain and prior swelling in the wrists and small joints of the hands and feet, which are commonly affected in rheumatoid arthritis.\\
4. previous history of joint pain and swelling: there is a history of pain and swelling in specific joints, particularly wrists and small joints of hands which is consistent with rheumatoid arthritis.\\
5. low-grade fever: the patient has a fever history under 100°f, which can occur in rheumatoid arthritis.\\
6. no skin rash: there is no presence of a skin rash, which aligns with rheumatoid arthritis as skin rashes often point toward other connective tissue disorders.}\\

\textcolor{violet}{\textit{Points against this diagnosis:}}\\
\texttt{1. pain increases with physical activity: while rheumatoid arthritis typically improves with activity, in this case, physical activity increases pain, which is less typical but can still occur.\\
2. warmth in joints: the presence of joint warmth might more commonly indicate active inflammation, but absence of definitive joint swelling is a potential contradiction.\\
3. absence of good painkiller response: only mild relief is obtained from painkillers, which may suggest atypical response, as good response to nsaids is often expected in rheumatoid arthritis.\\
4. axial and back pain: the involvement of the lower and upper back could suggest axial involvement, which is less typical in rheumatoid arthritis compared to other kinds of arthritis.}\\

\textcolor{violet}{\textit{Explanation for considering the diagnosis:}}\\
\texttt{despite some atypical symptoms, the presence of early morning stiffness, localized joint pain and swelling at the wrists and small joints of the hands, the pattern of symptoms over time, and fever support a diagnosis of rheumatoid arthritis. \hl{the absence of deforming swelling in other joints or additional systemic features typically rules out broader connective tissue disorders.}}\\

\textcolor{violet}{\textit{Other possible differentials:}}\\
\texttt{1. psoriatic arthritis: given the joint pain, swelling, and involvement of both peripheral and possibly axial joints, psoriatic arthritis could be considered despite the lack of skin psoriasis.\\
2. osteoarthritis: although typically not associated with early morning stiffness, the worsening of pain with activity and rest could potentially indicate osteoarthritis, requiring consideration of other influencing factors like age and the presence of joint degeneration.\\
3. ankylosing spondylitis: axial arthritis and back pain might indicate ankylosing spondylitis, though the lack of typical radiographic evidence and other symptoms might reduce its likelihood.\\
4. fibromyalgia: given chronic pain, although with fewer inflammatory signs, fibromyalgia may present similarly, especially with atypical joint pain response to physical activity and rest.\\
5. systemic lupus erythematosus (sle): while less likely due to the absence of rash, lupus should still be considered due to the joint involvement and systemic nature of symptoms.}\\

\end{tiny}
\end{multicols}
\end{tcolorbox}

\end{document}